\documentclass[letterpaper, 10 pt, conference]{ieeeconf}  % Comment this line out if you need a4paper

\IEEEoverridecommandlockouts                              % This command is only needed if 
\usepackage{amsmath,amssymb,amsfonts}
\usepackage{algorithmic}
\usepackage{graphicx}
\usepackage{textcomp}
\usepackage{xcolor}
\usepackage{booktabs}
\usepackage{array}
\usepackage{balance}
\usepackage{cite}

\title{\LARGE \bf
GPO: Global Plane Optimization for Fast and Accurate \\ 
Monocular SLAM Initialization
}

\author{Sicong Du$^{\dagger\ddagger1,2}$, Hengkai Guo$^{\dagger3}$, Yao Chen$^{3}$, Yilun Lin$^{1,2}$, Xiangbing Meng$^{*1,2}$, Linfu Wen$^{3}$, Fei-Yue Wang$^{1,2}$% <-this % stops a space
\thanks{This work was supported by National Natural Science Foundation of China 61533019, Beijing Natural Science Foundation 8172018 and Intel Collaborative Research Institute for Intelligent and Automated Connected Vehicles.}% <-this % stops a space
\thanks{$^{\dagger}$Joint first authors}
\thanks{$^{*}$Corresponding author}
\thanks{$^{\ddagger}$This work was done during Sicong Du's internship at ByteDance.}
\thanks{$^{1,2}$Sicong Du, Yilun Lin, Xiangbing Meng and Fei-Yue Wang are with the PA$^{2}$R$^{2}$T Group, the State Key Laboratory of Complex System Management and Control, Institute of Automation, Chinese Academy of Sciences, Beijing 100190, China. They are also with the University of Chinese Academy of Sciences, Beijing, 10049, China. {\tt\small \{dusicong2018, yilun.lin, xiangbing.meng, feiyue.wang\} @ia.ac.cn}}% 
\thanks{$^{3}$Hengkai Guo, Yao Chen and Linfu Wen are with the ByteDance AI Lab, Beijing 100089, China. {\tt\small \{guohengkai, chenyao.alejandro, wenlinfu\} @bytedance.com}}%
}

\begin{document}

\maketitle
\thispagestyle{empty}
\pagestyle{empty}

%%%%%%%%%%%%%%%%%%%%%%%%%%%%%%%%%%%%%%%%%%%%%%%%%%%%%%%%%%%%%%%%%%%%%%%%%%%%%%%%
\begin{abstract}
Initialization is essential to monocular Simultaneous Localization and Mapping (SLAM) problems. This paper focuses on a novel initialization method for monocular SLAM based on planar features. The algorithm starts by homography estimation in a sliding window. It then proceeds to a global plane optimization (GPO) to obtain camera poses and the plane normal. 3D points can be recovered using planar constraints without triangulation. The proposed method fully exploits the plane information from multiple frames and avoids the ambiguities in homography decomposition. We validate our algorithm on the collected chessboard dataset against baseline implementations and present extensive analysis. Experimental results show that our method outperforms the fine-tuned baselines in both accuracy and real-time.
\end{abstract}

%%%%%%%%%%%%%%%%%%%%%%%%%%%%%%%%%%%%%%%%%%%%%%%%%%%%%%%%%%%%%%%%%%%%%%%%%%%%%%%%
\section{INTRODUCTION}

Monocular Simultaneous Localization and Mapping (SLAM) aims to concurrently estimate the camera trajectory and reconstruct the unknown environment from a single input video. It has been widely used in the field of augmented reality (AR) \cite{billinghurst_survey_2015, carmigniani2011augmented} and autonomous driving \cite{ros_visual_2012, chen2018parallel, kebria2019deep}. Initialization is usually mandatory to bootstrap a monocular SLAM system. During the initialization, camera poses and an initial map are built for the subsequent tracking and mapping. A poor initialization slows down the convergence of the system or even leads to localization failures. 

General initialization methods for monocular SLAM are based on fundamental matrix decomposition \cite{mur-artal_orb-slam:_2015, qin_vins-mono:_2018, qin_robust_2017}. A 3D map can then be obtained by triangulation. Afterwards, a Perspective-n-Point(PnP) \cite{lepetit_epnp:_2009} method is performed to estimate poses of other frames. The whole process can be followed by incremental and local structure-from-motion (SfM) \cite{snavely_photo_2006, cui_global_2015, ozyesil_robust_2015, sun2017geographic, wilson_robust_2014}.

However, such initialization mechanisms have some drawbacks. Firstly, a sufficiently large parallax is required for an accurate triangulation and feature point depth can not be estimated in pure rotation. Secondly, due to the large scale of the SfM problem, it takes a long time to converge \cite{triggs_bundle_1999}. Lastly, man-made scenes often consist of planar structures such as floors and walls \cite{zhou_robust_2012, chen2019progressive, kim2019multi}, which leads to degeneration of the fundamental matrix.

In order to handle the planar regularity, some methods estimate the homography \cite{forster_svo:_2014, mur-artal_orb-slam:_2015} between two frames instead of the fundamental matrix. But the geometric ambiguity in homography decomposition makes it tricky to design good selection strategies suited for different applications \cite{mur-artal_orb-slam:_2015}. Moreover, only two-frame observations are used for all the systems above. If we can take the advantage of more frames for a planar scene, it can be expected to have more accurate estimation.

To address all these problems, we propose a fast and accurate initialization method for monocular SLAM. In our method, we first estimate homographies in a sliding window between the first frame and the current frame with RANSAC \cite{fischler_random_1981}. Then we propose global plane optimization (GPO) to minimize the 2D reprojection error of corresponding points with respect to the plane normal and scaled translations. Finally we estimate the 3D points on the plane using the plane equation. The core of our method is to avoid the homography decomposition by using the information in all frames. We also reconstruct the planar map without triangulation. Moreover, by reducing the number of variables in the optimization, our algorithm achieves significant real-time improvements.

The proposed methods are evaluated experimentally on collected chessboard dataset with trajectory and plane metrics. We implement several strong multi-frame baselines, including aggregation-based and optimization-based methods. We show that the proposed GPO outperforms other initialization methods in both accuracy and real-time.

To this end, we summarize our contributions as follow: 
\begin{itemize}
\item We develop a novel initialization method for monocular SLAM, which, to the best of our knowledge, is the first SLAM initialization method that fully makes use of multi-frame planar information.
\item We propose several initialization baselines and conduct exhaustive experiments to validate our method.
\item We propose novel evaluation metrics on the accuracy of plane estimation, so as to resolve the limitations of absolute translation error criterion. 
\end{itemize}

The rest of the paper is structured as follows. In Sec.\uppercase\expandafter{\romannumeral2}, we review the literature in related fields. In Sec.\uppercase\expandafter{\romannumeral3}, we present an overview of our method. Implementation details and experimental results are shown in Sec.\uppercase\expandafter{\romannumeral4}. Finally, the paper is concluded with a discussion and possible future work in Sect.\uppercase\expandafter{\romannumeral5}.

\section{RELATED WORK}

There is a large number of recent studies on monocular SLAM \cite{younes_keyframe-based_2017, taketomi_visual_2017, saputra_visual_2018, fuentes2015visual}. They can be classified into two categories: optimization-based methods and filtering-based methods. Filtering-based methods usually run faster because they marginalize historical states out recursively. However, they may be sub-optimal. Optimization-based methods can achieve better accuracy but the computational complexity is higher due to its iterative nature. In general, the performance of both SLAM frameworks rely heavily on the accuracy of the initial values \cite{qin_robust_2017}. 

Planar attributes in the initialization of monocular SLAM are studied as well. Forster et al. \cite{forster_svo:_2014} and Klein et al. \cite{klein_parallel_2007} assume that the scene is planar during initialization and use a homography to represent the transformation. The camera motion and the plane normal can be obtained by decomposing the homography matrix as described in \cite{faugeras_motion_1988, zhang_3d_1996, malis_deeper_2007}. The decomposition method in \cite{malis_deeper_2007} is analytical and the others are based on SVD. 

For more general scenes, a fundamental matrix is employed in \cite{mur-artal_orb-slam:_2015, qin_vins-mono:_2018, qin_robust_2017}. For example, ORB-SLAM \cite{mur-artal_orb-slam:_2015} decomposes the concurrently calculated homography and fundamental matrix into camera motions and generates several candidate models. The best model is selected according to a set of rigorous conditions. Hence, this process can take a long time dealing with complex scenes and camera motions. VINS \cite{qin_vins-mono:_2018} uses the five-point method \cite{nister_efficient_2004} to recover the camera pose by decomposing the fundamental matrix. After triangulation and PnP, a full bundle adjustment (BA) is performed to refine the initial poses and landmarks \cite{mur-artal_orb-slam:_2015, qin_vins-mono:_2018}. However, the triangulation requires sufficiently large parallaxes. 

% In addition, some direct methods \cite{engel_lsd-slam:_2014}\cite{engel_direct_2017} randomly initialize 3D points with large variance. Their accuracy are sometimes poor according to \cite{mur-artal_orb-slam:_2015} when the initial value is far from convergence point. 

Planar attributes can also be used in RGB-D SLAM system for extra structure constrains \cite{hsiao_keyframe-based_2017, ma_cpa-slam:_2016, kim_linear_2018, le_dense_2017, wang_submap_2018}. For example, the authors of \cite{hsiao_keyframe-based_2017} develop a keyframe-based dense planar SLAM to reconstruct large indoor environments using a RGB-D sensor, which can improve the plane extraction by generating a local depth map. Some monocular SLAM algorithms introduce the planes to enhance the performance \cite{servant_improving_2010} or deal with extreme cases \cite{yang_pop-up_2016} such as low-texture scenes. Zhou et al. \cite{zhou_robust_2012} show how to detect and track multiple planes in an uncalibrated video sequence \cite{wang_efficient_2005-1, he2019image, wang_simple_2004} in existence of dynamic outliers. Similar to our methods, Habbecke et al. \cite{habbecke2006iterative, habbecke2007surface} also leverage multi-frame homography optimization to enhance multi-view stereo reconstruction. But they choose to minimize the sum of squared differences of image intensities, in need of manual initialization of planes and larger computation cost.

%Vision-only SfM methods can calculate relative poses using multi-frame in sliding window in initialization process \cite{qin_vins-mono:_2018}. It is often solved by advanced libraries such as G2O \cite{kummerle_g_2011} and Ceres \cite{agarwal_ceres_2012}. The optimization in our method is similar with SfM. However, we decrease the variables by removing the map point positions to acclerate its convergence. The variables in our method only include plane normal and translations. Afterwards, we can recover the map points using the plane equation with normal known.

\section{METHODOLOGY}

\begin{figure}[tb]
\centering
\framebox{\includegraphics[width=7cm,height=7cm]{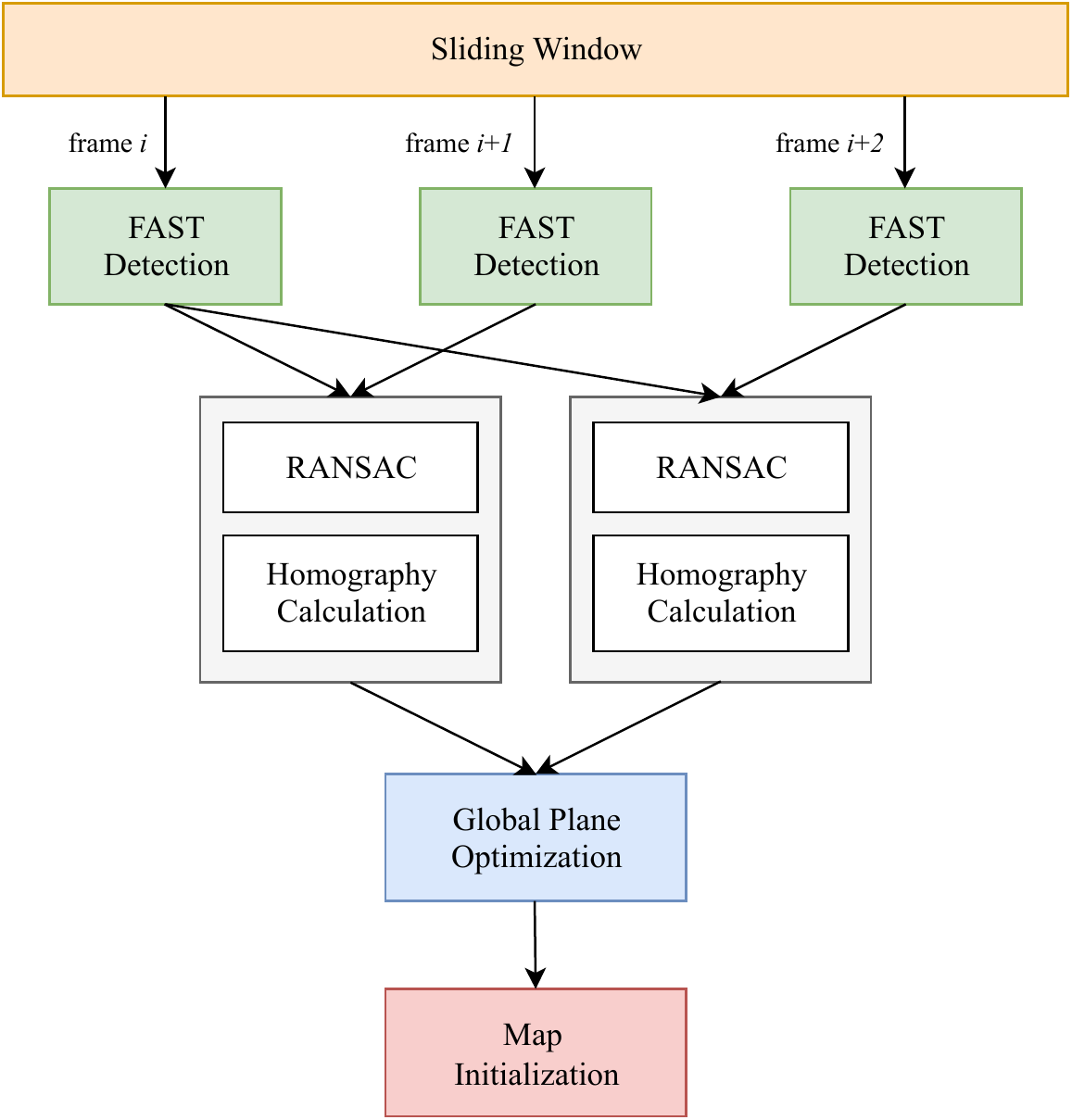}}
\caption{An illustration of the structure of our initialization method. We take 3 frames in the sliding window as an example. The system has 4 parts: feature points detection and tracking, homography calculation, global homography optimization and map initialization.}
\label{figsystem}
\end{figure}

The structure of the proposed initialization system is shown in Fig. \ref{figsystem}. There are 4 main parts: feature points detection and tracking, homography calculation using RANSAC \cite{fischler_random_1981}, global homography optimization and map initialization. The feature points are extracted and tracked in the measurement preprocessing. Homographies are then estimated according to the feature correspondences between the first frame and the current frame in sliding window. We implement a global plane optimization (GPO) method to obtain the plane parameters and camera poses. Finally we can recover the map points from the plane without triangulation.

We now define notations and coordinate frames in this paper. $\textbf{K}$ is the intrinsic matrix of a pinhole camera. ${_{c_{2}}\textbf{R}_{c_{1}}}$ is the rotation matrix which rotates the coordinates from camera frame ${c_{1}}$ to ${c_{2}}$ where the number indexes the image timestamp. ${_{c_{2}}\textbf{H}_{c_{1}}}$ denotes the homography induced by a plane observed from ${c_{1}}$ to ${c_{2}}$. ${_{w}}\textbf{t}_{{c_{2}}{c_{1}}}$ is the translation of ${c_{1}}$ with respect to ${c_{2}}$, represented in the world frame ${w}$. We consider ${{_w}\textbf{n}}$ and ${_{c_{1}}\textbf{n}}$ as the plane normal represented in the world frame ${w}$ and camera frame ${c_{1}}$ respectively. ${d_{c_{1}}}$ is the distance from camera ${c_{1}}$ to the plane.
$\boldsymbol{\pi}(\cdot)$ is the normalization function defined by

\begin{equation}
[x/z,y/z]^{T} = \boldsymbol{\pi}([x,y,z]^T)
\end{equation}

\subsection{Fundamentals of Homographies}
We first introduce the fundamentals for homography estimation and decomposition.

\subsubsection{Homography Estimation}
The homography can be estimated with the feature correspondences from two frames. The homography ${_{c_{2}}\textbf{H}_{c_{1}}}$ is defined as follows:

\begin{equation}
_{c_{2}}\textbf{H}_{c_{1}}=\textbf{K}({_{c_{2}}\textbf{R}_{c_{1}}}-\frac{_{c_{2}}\textbf{t}_{{c_{2}}{c_{1}}}}{d_{c_{1}}}\cdot{_{c_{1}}\textbf{n}^{T}}){\textbf{K}^{-1}}
\end{equation}
where ${_{c_{2}}\textbf{t}_{c_{2}c_{1}}}={(t_{x},t_{y},t_{z})}^{T}$ and ${_{c_{1}}\textbf{n}}={(n_{x},n_{y},n_{z})}^{T}$. 4-point RANSAC is applied to remove outliers.

% In general, there are three homography decomposition methods: Faugeras, Zhang and Inria. The first two methods are based on the SVD decomposition. Inria avoids it and can obtain an analytical result with closed forms\cite{malis_deeper_2007}.
\subsubsection{Homography Decomposition}
We can obtain the camera pose and plane parameters by decomposing the estimated homography \cite{malis_deeper_2007}. However, conventional decomposition methods have three major drawbacks. Firstly, small perturbation to the homography can make the decomposition result deviate from the correct value. Secondly, the decomposition of the homography to obtain the normal is mostly erroneous due to the bilinear nature of the normal and the translation \cite{singhal_top_2014}. Thirdly, they return up to 4 solutions in most cases. Strategies should be designed to select the most reasonable solution \cite{mur-artal_orb-slam:_2015}. Some methods select the best solution of plane normal which maximizes consistency within multiple frames, but the noise from measurements and estimation makes it difficult to find the consistent normal. Therefore, we aim to design a more robust and accurate method to avoid the matrix decomposition.

\subsection{Global Plane Optimization}
\label{param}
\begin{figure}[tb]
\centerline{\includegraphics[width=7cm,height=3.5cm]{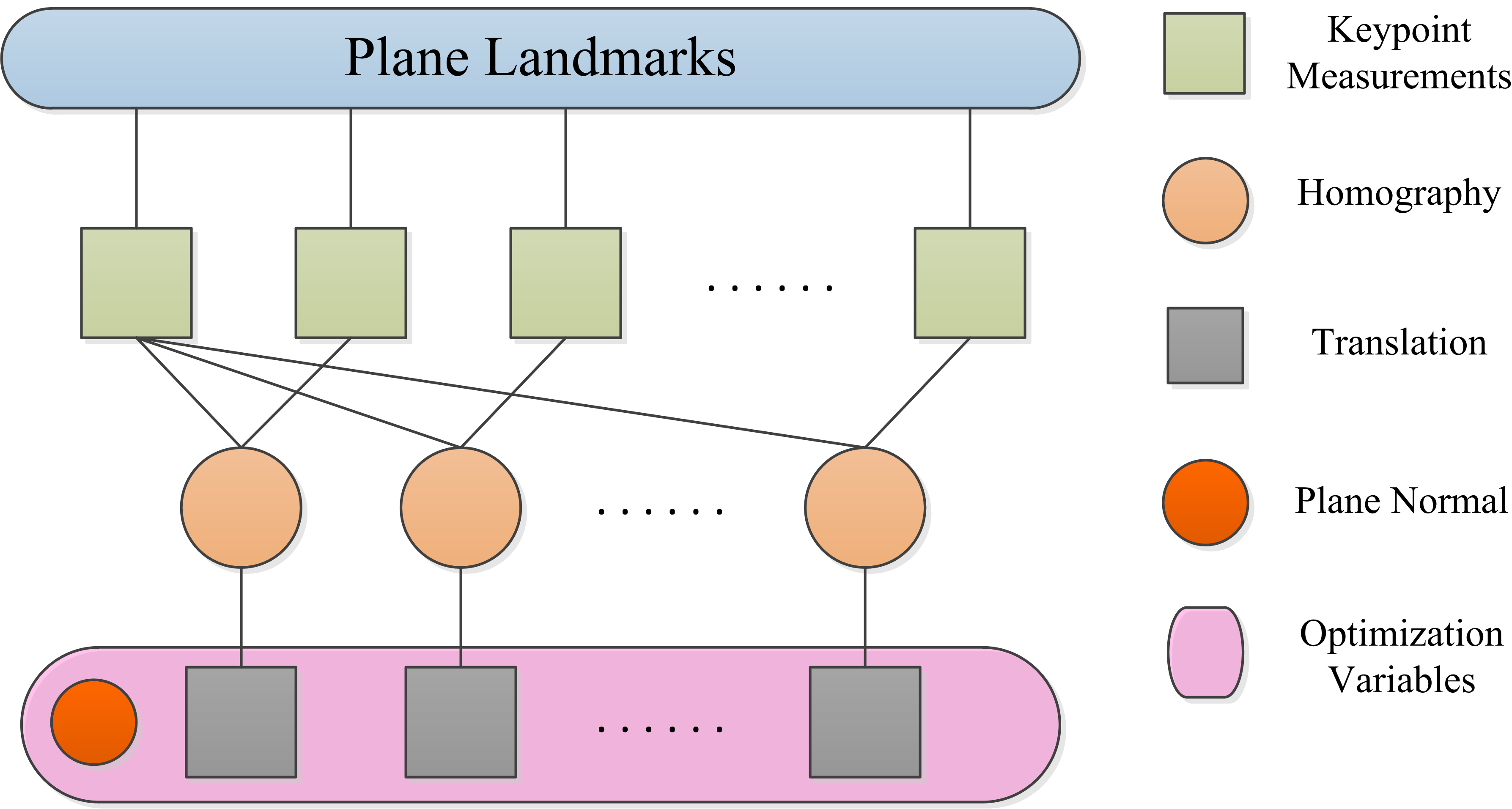}}
\caption{An illustration of optimization process. The observations are keypoint measurements. The unknowns are plane normal and camera translations included in the pink area.}
\label{figoptimization}
\end{figure}

In our method, we only leverage the homography estimation for outlier detection. To estimate the camera poses, we minimize the 2D reprojection error induced by the homography. To simplify the optimization, we first estimate the rotation for each frame. Rotation can be obtained directly using the methods in \cite{kneip_direct_2013, kneip_finding_2012, zhang_fast_2018}, or by integration of the angular rates from the gyroscope if available. Then the homography can be expressed as:
\begin{equation}
_{c_{2}}\textbf{H}_{c_{1}}=\textbf{K}\cdot{_{c_{2}}\textbf{R}_{w}}\cdot({\textbf{I}-\frac{_{w}\textbf{t}_{{c_{2}}{c_{1}}}}{d_{c_{1}}}\cdot{_{w}\textbf{n}^{T}}})\cdot{_{w}\textbf{R}_{c_{1}}}\cdot{\textbf{K}^{-1}}
\end{equation}

The variables are the plane normal ${_{w}\textbf{n}}$ and the translations ${_{w}\textbf{t}_{{c_{i}}{c_{1}}}}$ as illustrated in Fig. \ref{figoptimization}. Supposing that we have $m$ pairs of adjacent frames, the variables can be defined as:
\begin{equation}
\boldsymbol{\Omega}=[{_{w}\textbf{n}},{_{w}\textbf{t}_{{c_{2}}{c_{1}}}},{_{w}\textbf{t}_{{c_{3}}{c_{1}}}},\cdots,{_{w}\textbf{t}_{{c_{m}}{c_{1}}}}]
\end{equation}

We consider each frame ${c_{i}}$ has $n_i$ pairs of feature points selected. The optimization objective is 
\begin{equation}
\mathop{\arg\min}_{\boldsymbol{\Omega}}\sum\limits_{i=2}^m\sum\limits_{j=1}^{n_i} {\|\textbf{p}_{{c_{i}}}^{j}-\boldsymbol{\pi}({_{{c_{i}}}}{\textbf{H}}{_{{c_{1}}}}\cdot{\textbf{p}_{{c_{1}}}^{j}})\|}_{2}
\end{equation}
where ${\textbf{p}_{{c_{i}}}^{j}}$ are the homogeneous coordinates of $j$-th point on ${c_{i}}$ image. The translations can be known only up to scale since the depth is unknown. Consequently, the camera to plane distance is absorbed by translations. Because our state does not contain the 3D points, we simply use zero translations and a plane normal parallel to the z axis of camera frame as initial values for optimization.

Note that the proposed GPO is more efficient than the bundle adjustment problem \cite{triggs_bundle_1999} due to the smaller problem size. It contains $(3m-1)$ parameters\footnote{Assume that the rotation are fixed and all frames contain the same number of points.} and $(2mn-2n)$ residuals, while the bundle adjustment contains $(3m+3n-3)$ parameters and $2mn$ residuals, where $m$ is the number of camera poses and $n$ is the number of landmarks. And the Schur complement trick \cite{triggs_bundle_1999} can also be applied on the translations to speed up the optimization.

%The total variable dimension is $(2+3m)$, which is less than the general BA. 

% The Jaccobian can be expressed as:
% \begin{equation}
% \frac{\partial \sum\limits_{i=2}^m\sum\limits_{j=1}^{n_i} {\|p_{c_{i}}^{j}-\pi({_{c_{i}}}H{_{c_{1}}}\cdot{p_{{c_{i}}}^{j}})\|}_{2}} {\partial n_{local}}
% \end{equation}
% where $n_{local}$ is the expression of normal in the process of local parameterization. According to the chain rule in derivative, it can be represented as:
% \begin{equation}
% \frac{\partial \sum\limits_{i=2}^m\sum\limits_{j=1}^{n_i} {\|p_{{c_{i}}}^{j}-\pi({_{{c_{i}}}}H{_{{c_{1}}}}\cdot{p_{{c_{1}}}^{j}})\|}_{2}} {\partial ({_{{c_{i}}}}H{_{c_{i}}}\cdot{p_{1}^{j}})}\cdot \frac{\partial ({_{{c_{i}}}}H{_{{c_{1}}}}\cdot{p_{1}^{j}})} {\partial n_{global}} \cdot \frac{\partial n_{global}} {\partial n_{local}}
% \end{equation}
% The first two terms can be solved using matrix derivatives. The third term's solution method is referenced from \cite{bloesch_iterated_2017}.
% \begin{equation}
% \frac{\partial n_{global}} {\partial n_{local}}=n(\mu)^{\times}N(\mu)
% \end{equation}
% \begin{equation}
% n(\mu):=\mu(e_{z})\in S^{2} \subset R^{3}
% \end{equation}
% \begin{equation}
% N(\mu):=[\mu(e_{x}),\mu(e_{y})] \in \mathbb(R)^{3\times2}
% \end{equation}
% where $e_{x}, e_{y}, e_{z}$ $\in$ $R^{3}$. They are the basis vectors for an orthonormal coordinates. $n(\mu)$ is the actual unit vector which represents rotating $e_{z}$ by $\mu$. $N(\mu)$ includes the rotated $e_{x}$ and $e_{y}$. Afterwards we can obtain the plane normal and translations between frames.

\subsection{Map Points Reconstruction}

After solving for the plane normal, we can construct the plane equation using the following expression:
\begin{equation}
\textbf{n}^T\textbf{P}+d=0
\end{equation}
where $\textbf{P}$ represents the 3D coordinates of feature points on this plane in the world coordinate frame. $d$ is the distance from camera to the plane up to scale. According to the camera projection model, we can get the following equation:
\begin{equation}
\textbf{K}({_{c_i}}\textbf{R}_{w}\cdot{\textbf{P}_{w}^{j}}+_{c_{i}}\textbf{t}_{{c_{i}}{w}})=m_{c_i}^{j}\cdot{\textbf{p}_{c_{i}}^{j}}
\end{equation}
where $m_{c_i}^{j}$ represents the depth of $j$-th point in the frame $ci$. After transformation, the depth can be calculated as:
\begin{equation}
m_{c_i}^{j}=\frac{-d+{\textbf{n}^T}_w\textbf{t}_{{c_{i}}{w}}}{\textbf{n}^T\cdot{_w}\textbf{R}_{c_i}\cdot{\textbf{K}^{-1}}{\textbf{p}_{{c_{i}}}^{j}}}
\end{equation}
We can use the above method to calculate the depth of each feature point in all frames and then average the depths for a single landmark.

\section{Experimental Results}
In this section, we validate our algorithm with our collected dataset. We first explain the dataset and evaluation metrics in our experiments. Then several strong baselines for multiple-frame initialization on plane scenes are proposed for comparison. After that, we verify the effectiveness of our RANSAC component in the GPO method. Finally, the proposed GPO algorithm is compared with all the baselines both on accuracy and speed. All the experiments are run on a MacBook Pro with 2.5GHz i7 CPU and 16GB memory.

\subsection{Dataset and Evaluation Metrics}
\begin{figure}[tb]
% \begin{minipage}{0.48\linewidth}
%   \centerline{\includegraphics[width=4.4cm,height=3.3cm]{c3.jpg}}
%   \footnotesize\centerline{(a)}
% \end{minipage}
% \hfill
% \begin{minipage}{0.48\linewidth}
%   \centerline{\includegraphics[width=4.4cm,height=3.3cm]{c4.jpg}}
%   \footnotesize\centerline{(b)}
% \end{minipage}
% \vfill
\begin{minipage}{0.48\linewidth}
  \centerline{\includegraphics[width=4.4cm,height=3.3cm]{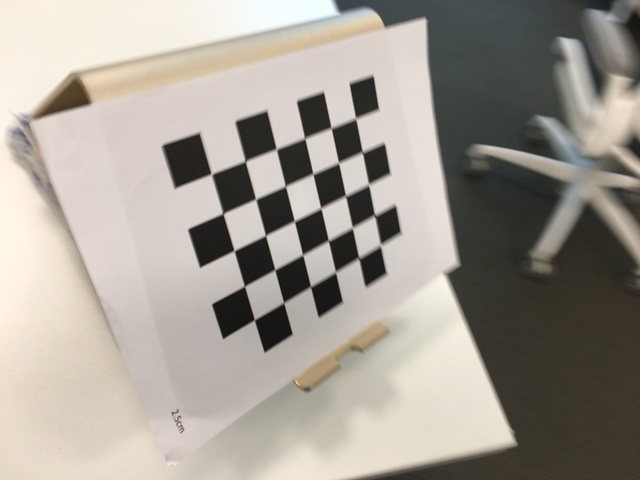}}
  %\footnotesize\centerline{(c)}
\end{minipage}
\hfill
\begin{minipage}{0.48\linewidth}
  \centerline{\includegraphics[width=4.4cm,height=3.3cm]{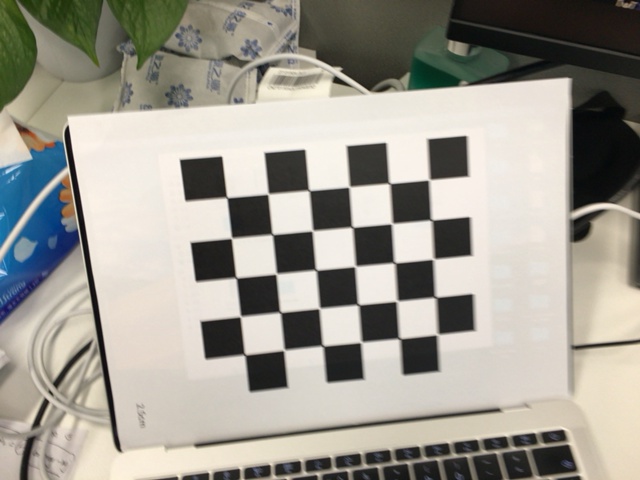}}
  %\footnotesize\centerline{(d)}
\end{minipage}
\caption{Sample images of our chessboard dataset.}
\label{samples}
\end{figure}

Our dataset consists of 9 video sequences captured by an iPhone 6s with different movement patterns. For each sequence, a chessboard is
viewed from different camera poses. We can obtain the ground truth trajectory of the camera and the plane parameters using PnP algorithm based on the known chessboard rig. Some sample images are shown in Fig. \ref{samples}.

For evaluation, we estimate and apply the similarity transformation $U$ to align our trajectory with the ground truth one using Umeyama algorithm \cite{umeyama_least-squares_1991}:
\begin{equation}
U=\mathop{\arg\min}_{s,\textbf{R},\textbf{T}}\sum\limits_{t=0}^{N-1}{\|(\textbf{Rp}[t]+\textbf{T})-s\textbf{p}_{GT}[t]\|}^{2}
\end{equation}
where $\textbf{p}$ and $\textbf{p}_{GT}$ are the camera positions of our methods and ground truth respectively and $s, \textbf{R}, \textbf{T}$ are the scale, rotation and translation of the similarity transformation, respectively.

We use three criterions to test the performance of our initialization methods. The absolute translation error (ATE) is defined by:

\begin{equation}
\epsilon_{ATE} = \sqrt{{\frac{1}{m}}\sum\limits_{i=1}^m{\|\textbf{p}^{'}[i]-\textbf{p}_{GT}[i]\|^{2}}}
\end{equation}
where $\textbf{p}^{'}$ is the aligned camera position.

The ATE metric can only measure the accuracy for estimated poses. For initialization, the quality of the point cloud is also important because we need it to localize the camera for the subsequent frame. We first estimate the plane using 3-point RANSAC. Then we propose two metrics for plane estimation: plane normal error (PNE) and plane distance error (PDE):
\begin{equation}
\epsilon_{PNE} = \arccos{(\textbf{n}^T\textbf{n}_{GT})}
\end{equation}
\begin{equation}
\epsilon_{PDE} = |d - d_{GT}|
\end{equation}
where $\textbf{n}^{T}\textbf{p} + d = 0$ and $\textbf{n}^{T}_{GT}\textbf{p} + d_{GT} = 0$ are the plane equations of our methods and ground truth respectively.

\subsection{Multi-frame Initialization Baselines for Plane Scenes}
\begin{figure}[tb]
\centerline{\begin{minipage}{0.9\linewidth}
  \centerline{\includegraphics[width=1.0\linewidth]{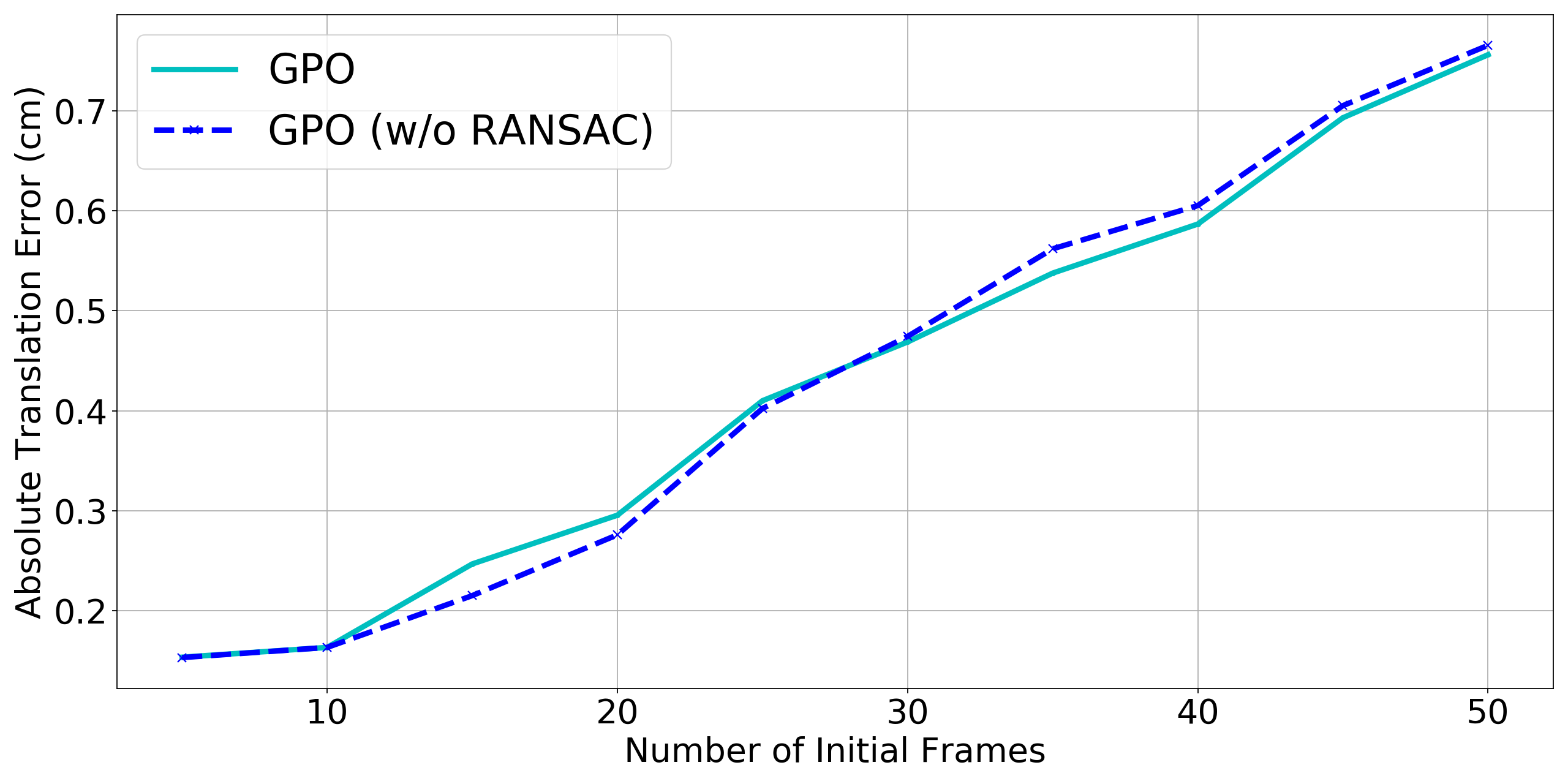}}
  \footnotesize\centerline{(a)}
\end{minipage}}
\vfill
\centerline{\begin{minipage}{0.9\linewidth}
  \centerline{\includegraphics[width=1.0\linewidth]{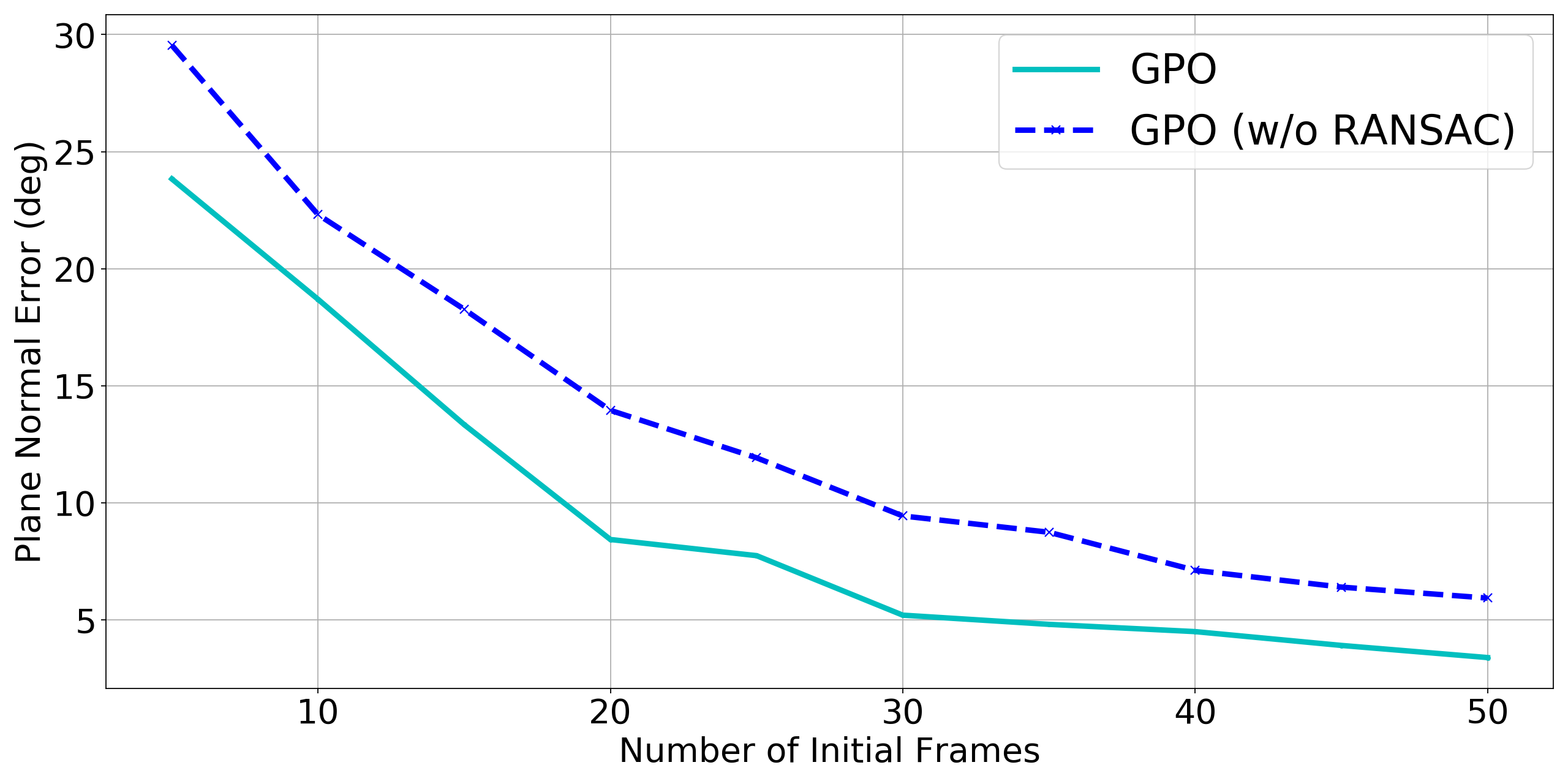}}
  \footnotesize\centerline{(b)}
\end{minipage}}
\vfill
\centerline{\begin{minipage}{0.9\linewidth}
  \centerline{\includegraphics[width=1.0\linewidth]{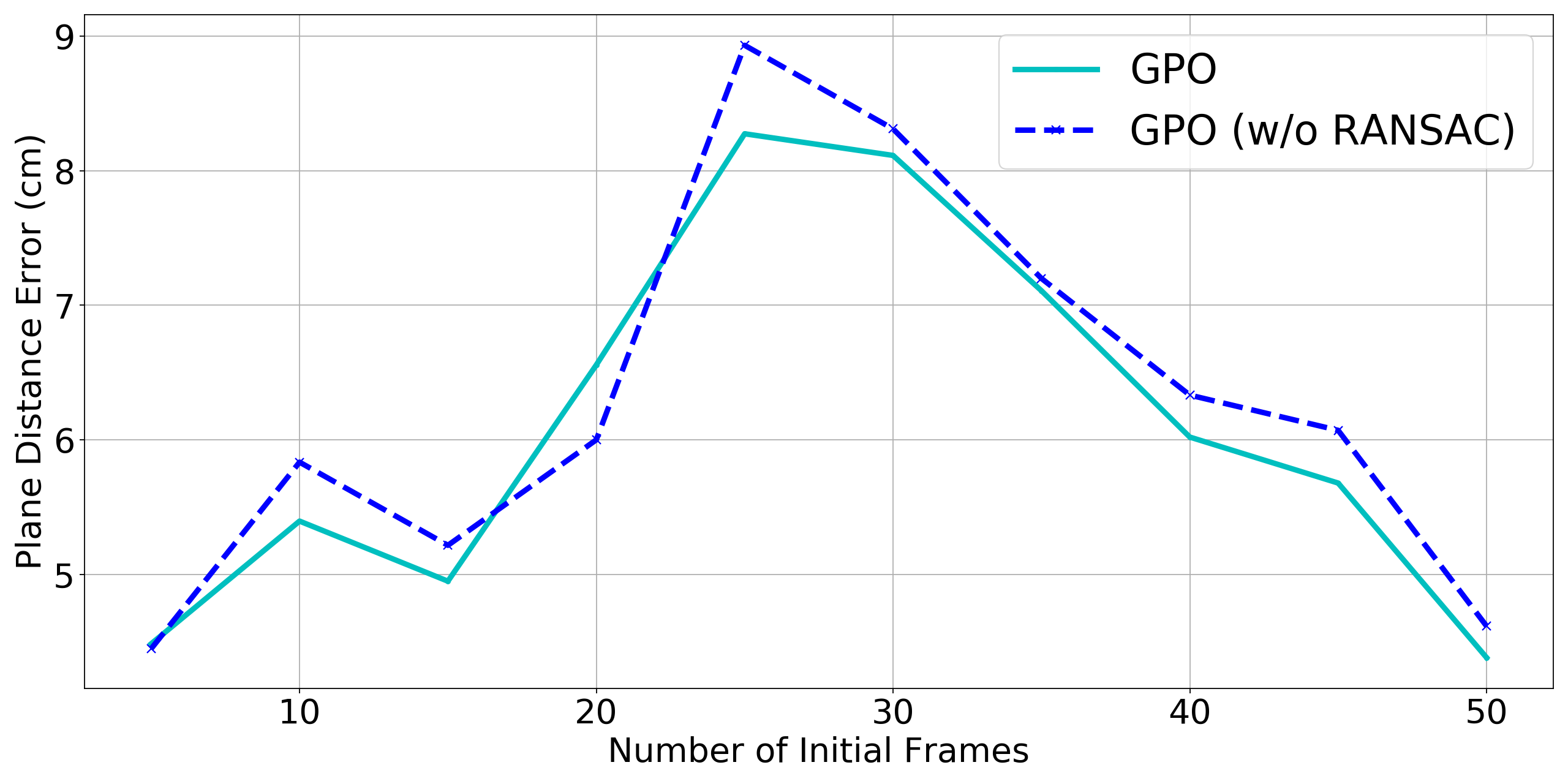}}
  \footnotesize\centerline{(c)}
\end{minipage}}
\caption{ATE, PNE and PDE of our GPO method with or without RANSAC.}
\label{fig_ransac}
\end{figure}

To the best of our knowledge, there are no existing methods for initialization in plane scene making use of multiple frame information. So we propose several strong baselines in this part to make the comparison as fair as possible. We group them into two categories: multi-frame aggregation and multi-frame optimization.

\subsubsection{Multi-frame Aggregation}
We consider two aggregation-based methods based on the results of two-frame homography decomposition. 

The first method is based on PnP \cite{lepetit_epnp:_2009}. The homography is decomposed from two frames. Then the initial point cloud can be obtained through triangulation. Afterwards it applies PnP and triangulation frame-by-frame to estimate the poses and points of the remaining frames.

The second method uses clustering method to calculate plane normal with multi-frame inspired by DBSCAN \cite{ester_density-based_1996}. Obviously all the candidate normal vectors are distributed on a surface of a unit sphere with the constraint of $n_{x}^{2} + n_{y}^{2} + n_{z}^{2} = 1$. We obtain the location on the surface with maximum density as the normal of the plane.

\subsubsection{Multi-frame Optimization}
We design three optimization algorithms: vanilla bundle adjustment (BA), fixed-plane bundle adjustment (FPBA) and plane bundle adjustment (PBA). All the methods require good initialization of poses and points, so we use DBSCAN for bootstrap. For BA, we also implement a version using PnP for initialization to make it similar to the vision part of VINS initialization method \cite{qin_vins-mono:_2018}.

The three different BA methods all minimize the reprojection error in terms of camera poses and landmarks. We consider each frame $i$ has $n_j, j=1, 2, \ldots, m$ pairs of feature points selected, the reprojection error can be expressed as follows:
\begin{equation}
\sum\limits_{i=1}^m\sum\limits_{j=1}^{n_i} {\|\boldsymbol{\pi}[\textbf{K}({_{c_i}}\textbf{R}{_{w}}\textbf{P}_{w}^{j}+{_{w}}\textbf{t}_{{c_{i}}{w}})] - \textbf{p}_{c_{i}}^{j}\|}_{2}
\end{equation}

For BA, the unknowns include landmarks and camera 
poses. PBA add a planar constraint that all the 3D map points are on the same plane to the optimization function. Hence a 3D point can be expressed using 2 variables with a known plane. As for FPBA, we fix the plane normal as constants during optimization process. For all the optimization methods without plane constraints, we skip the homography-based RANSAC during feature extraction.

%For BA, The unknowns include landmarks and camera poses. The dimension of the unknowns is $(3n+6m)$, where $n$ is the number of landmarks and $m$ is the number of camera poses.

%Different from BA, PBA add a planar constraint that all the 3D map points are on the same plane to the optimization function. Hence a 3D point can be expressed using 2 variables with a known plane. The dimension of the unknowns is $(2n+6m+2)$. The last constant $2$ is the freedom of the unit normal vector. 

%As for FPBA, we fix the plane normal as constants during optimization process. Therefore the number of variables is further decreased to $(2n+6m)$.

%The baseline method includes 3 steps. Firstly, the baseline approach select two frames and calculate the homography matrix and then decompose it. This part is different from VINS. VINS uses Five-point method to recover camera pose from fundamental matrix. According to the decomposition results, if the parallex is long enough and the number of feature points computed from triangulation is greater than the threshold, the baseline method uses PnP to obtain the camera pose. 

%Based on the the locations pf 3D points, the baseline approach implement PnP to other frames with corresponding 2D projections on the image known iteratively. Finally, a global Bundle Adjustment is applied to minimize the reprojection error to optimize the feature positions and camera pose.

\subsection{Implementation Details}
In all our experiments, we use FAST detector \cite{rosten_machine_2006} and KLT tracking\cite{baker_lucas-kanade_2004} as inputs. The rotation matrix is obtained from the integration of IMU raw data for all methods. We use Ceres \cite{ceres-solver} to implement all the optimizations. The trust region method is set to Dogleg and the number of maximum iterations is 300. The plane normal is parameterized by a unit quaternion as in \cite{bloesch_iterated_2017}.

\subsection{Ablation Study for RANSAC}
\label{gpo_ransac}
To demonstrate the necessity of RANSAC, we perform an ablation study with it. The results are shown in Fig. \ref{fig_ransac}. From the ATE, these two methods perform comparable. But for the plane metrics, GPO with RANSAC outperforms the one without RANSAC on plane normal estimation significantly and on plane distance estimation slightly. This is because points out of the planes will greatly harm the accuracy of plane estimation in a multi-plane scene.

\begin{figure}[tb]
\centerline{\begin{minipage}{0.9\linewidth}
  \centerline{\includegraphics[width=1.0\linewidth]{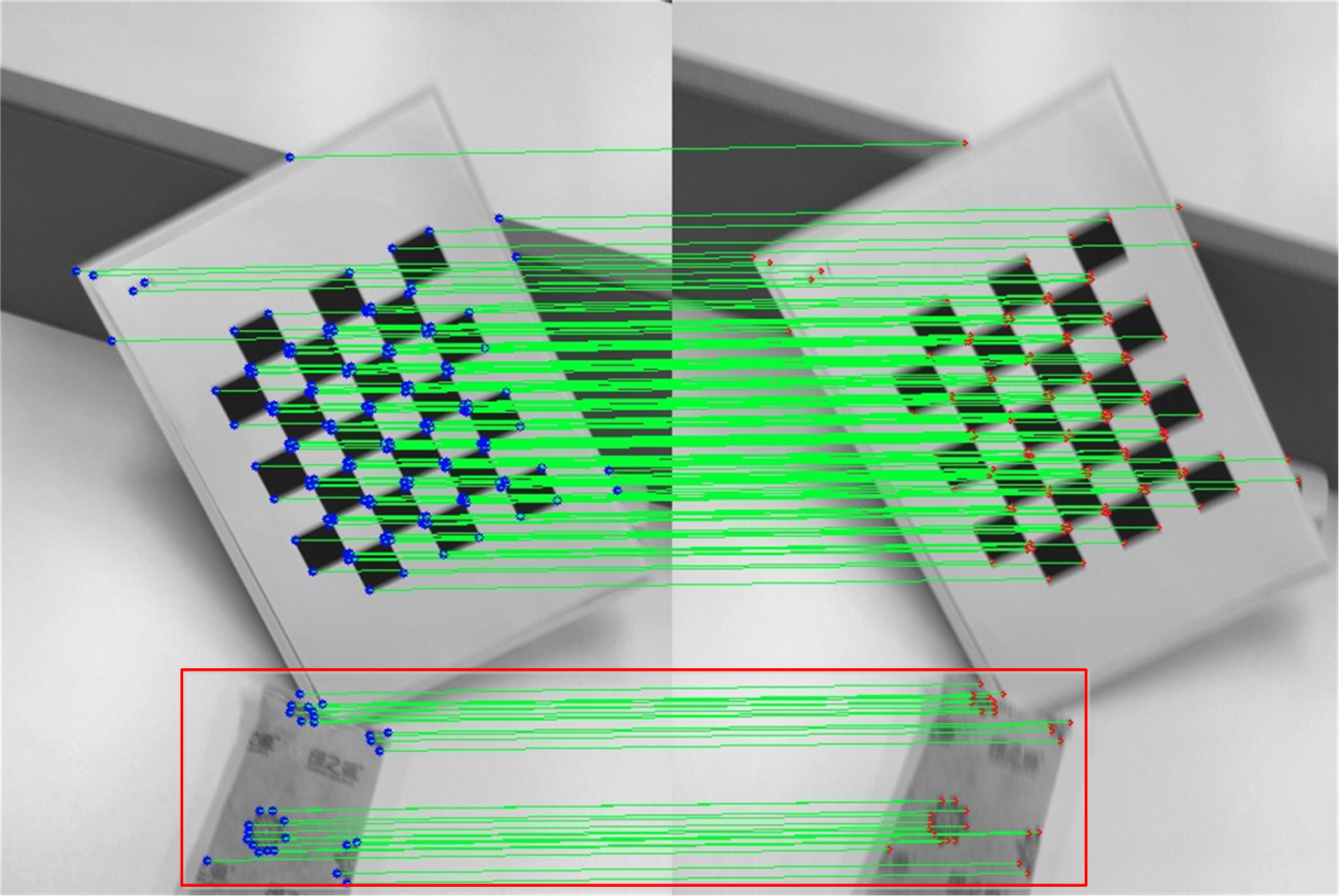}}
  \footnotesize\centerline{(a)}
\end{minipage}}
\vfill
\centerline{\begin{minipage}{0.9\linewidth}
  \centerline{\includegraphics[width=1.0\linewidth]{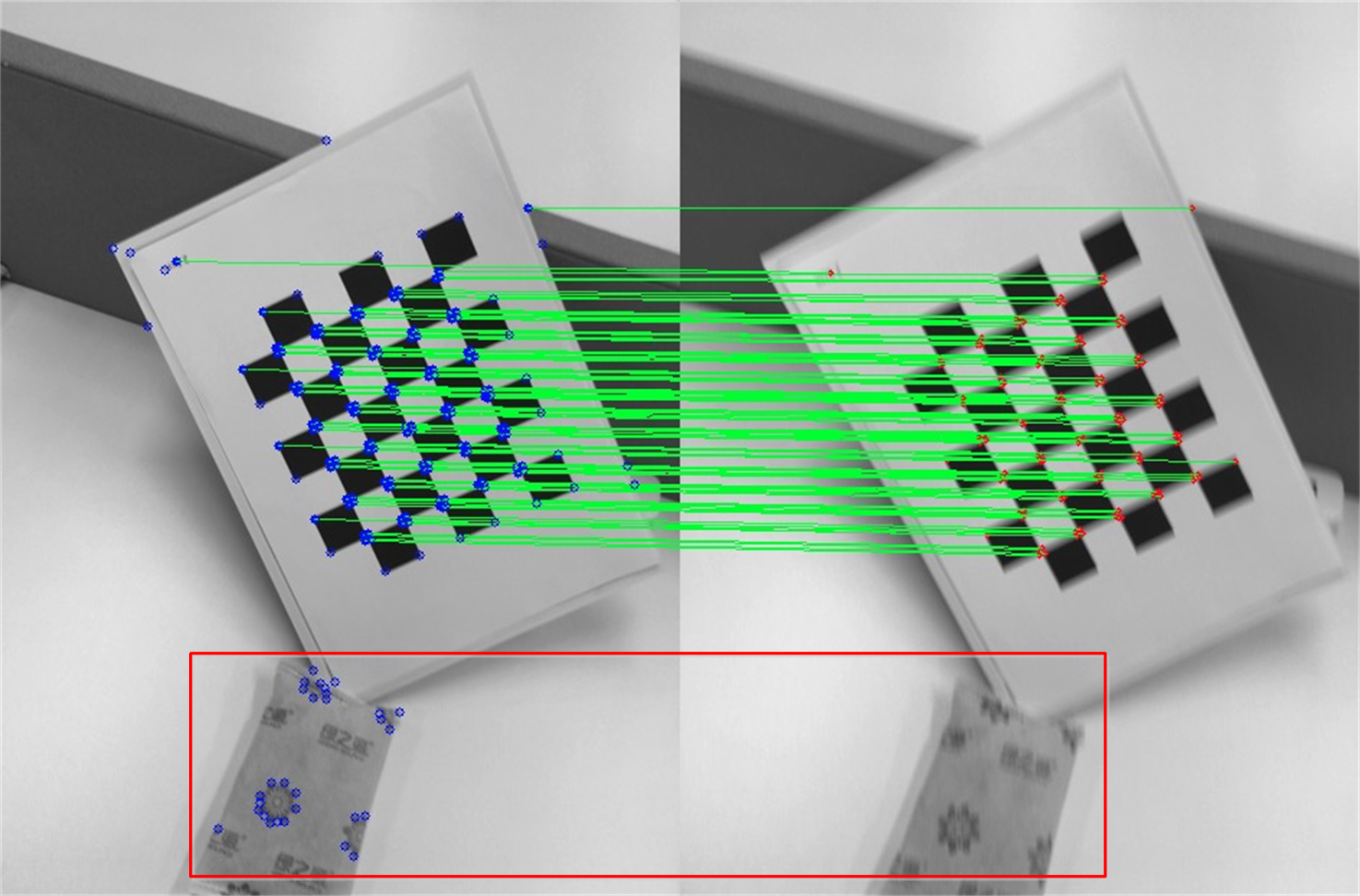}}
  \footnotesize\centerline{(b)}
\end{minipage}}
\caption{Examples of matched feature points (a) before, and (b) after RANSAC from our dataset.}
\label{fig_ransac_vis}
\end{figure}

We visualize some frames to take a closer look at such difference. Fig. \ref{fig_ransac_vis} shows the matched points before and after RANSAC. The points from the object on the desk is incorrectly included by the plane optimization when no RANSAC is applied.

\subsection{Performance Compared to Aggregation-based Methods}
\begin{figure}[tb]
\centerline{\begin{minipage}{0.9\linewidth}
  \centerline{\includegraphics[width=1.0\linewidth]{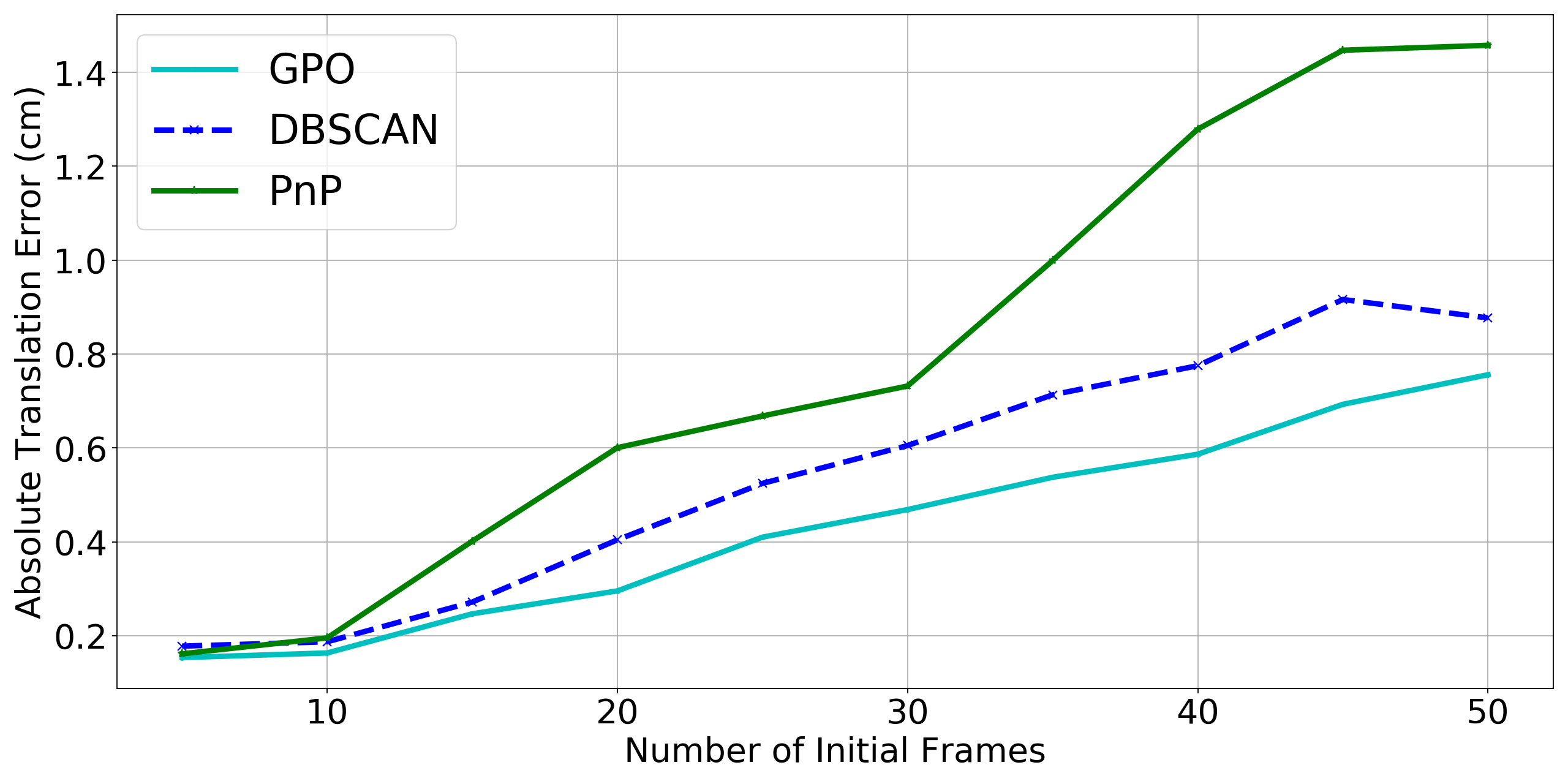}}
  \footnotesize\centerline{(a)}
\end{minipage}}
\vfill
\centerline{\begin{minipage}{0.9\linewidth}
  \centerline{\includegraphics[width=1.0\linewidth]{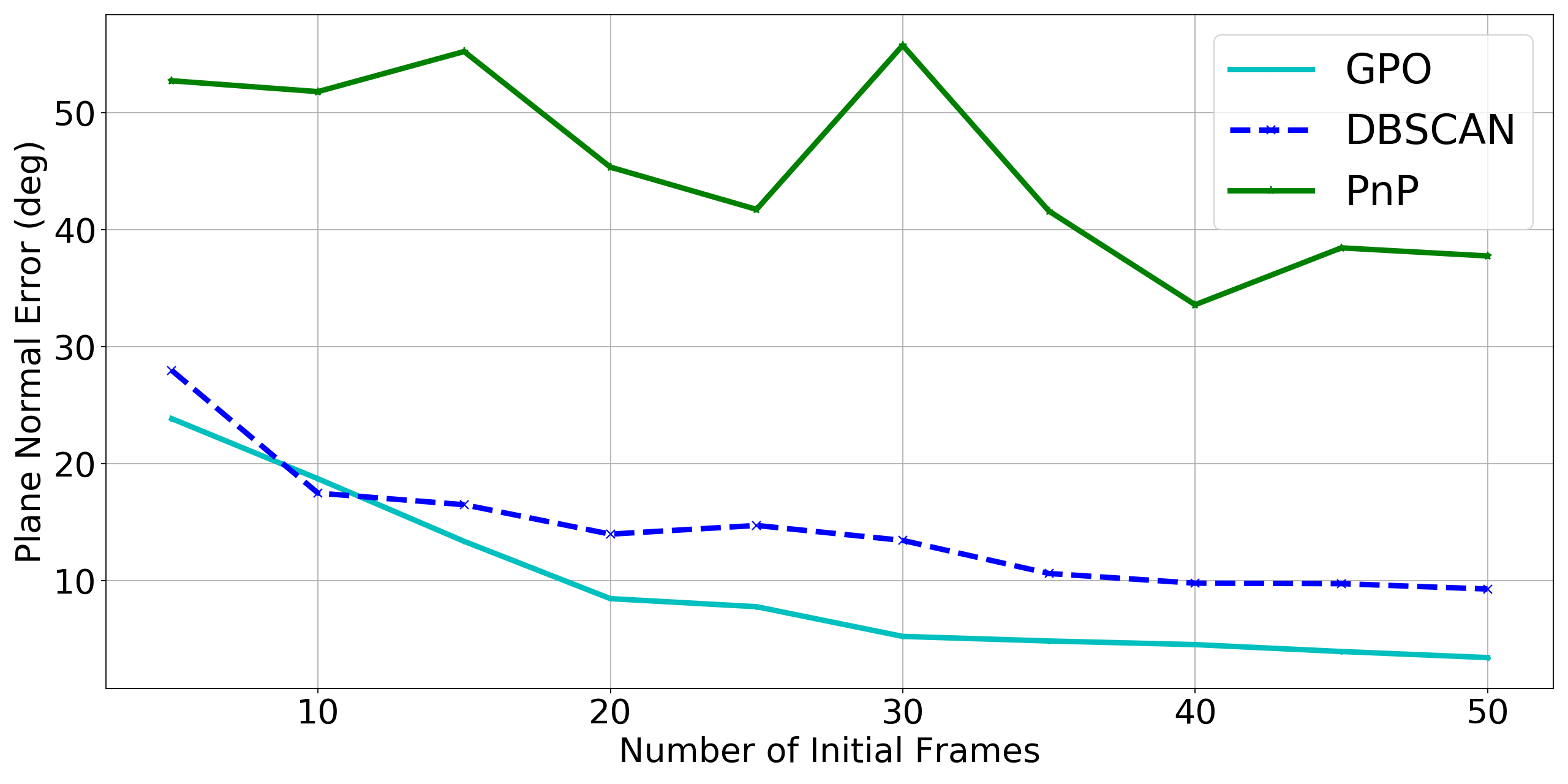}}
  \footnotesize\centerline{(b)}
\end{minipage}}
\vfill
\centerline{\begin{minipage}{0.9\linewidth}
  \centerline{\includegraphics[width=1.0\linewidth]{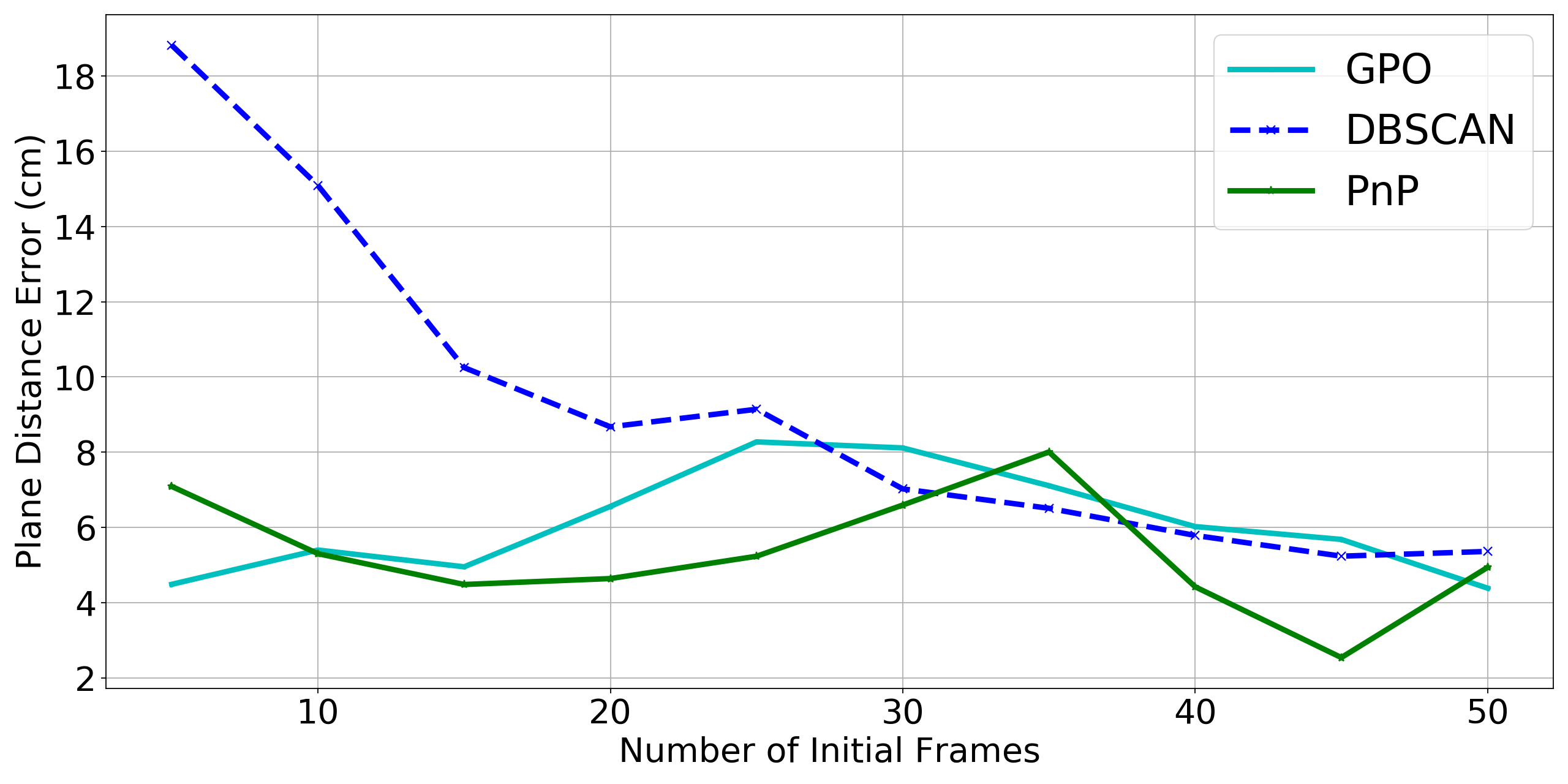}}
  \footnotesize\centerline{(c)}
\end{minipage}}
\caption{ATE, PNE and PDE of our GPO method and aggregation-based baselines.}
\label{fig_simple}
\end{figure}

We compare GPO to aggregation-based baseline methods, including PnP and DBSCAN in Fig. \ref{fig_simple}. Our GPO yields the best accuracy on both ATE and PNE among these algorithms and comparable accuracy on PDE with PnP, because it globally optimizes the poses and normal on all frames. PnP only employs two-frame homography, so the normal errors are much larger than the other methods. Besides, the initial errors of plane distance (5 to 25 frames) for DBSCAN are much larger. It indicates that more frames are necessary to achieve a good result for clustering-based methods.

\subsection{Performance Compared to Optimization-based Methods}
\begin{figure}[tb]
\centerline{\begin{minipage}{0.9\linewidth}
  \centerline{\includegraphics[width=1.0\linewidth]{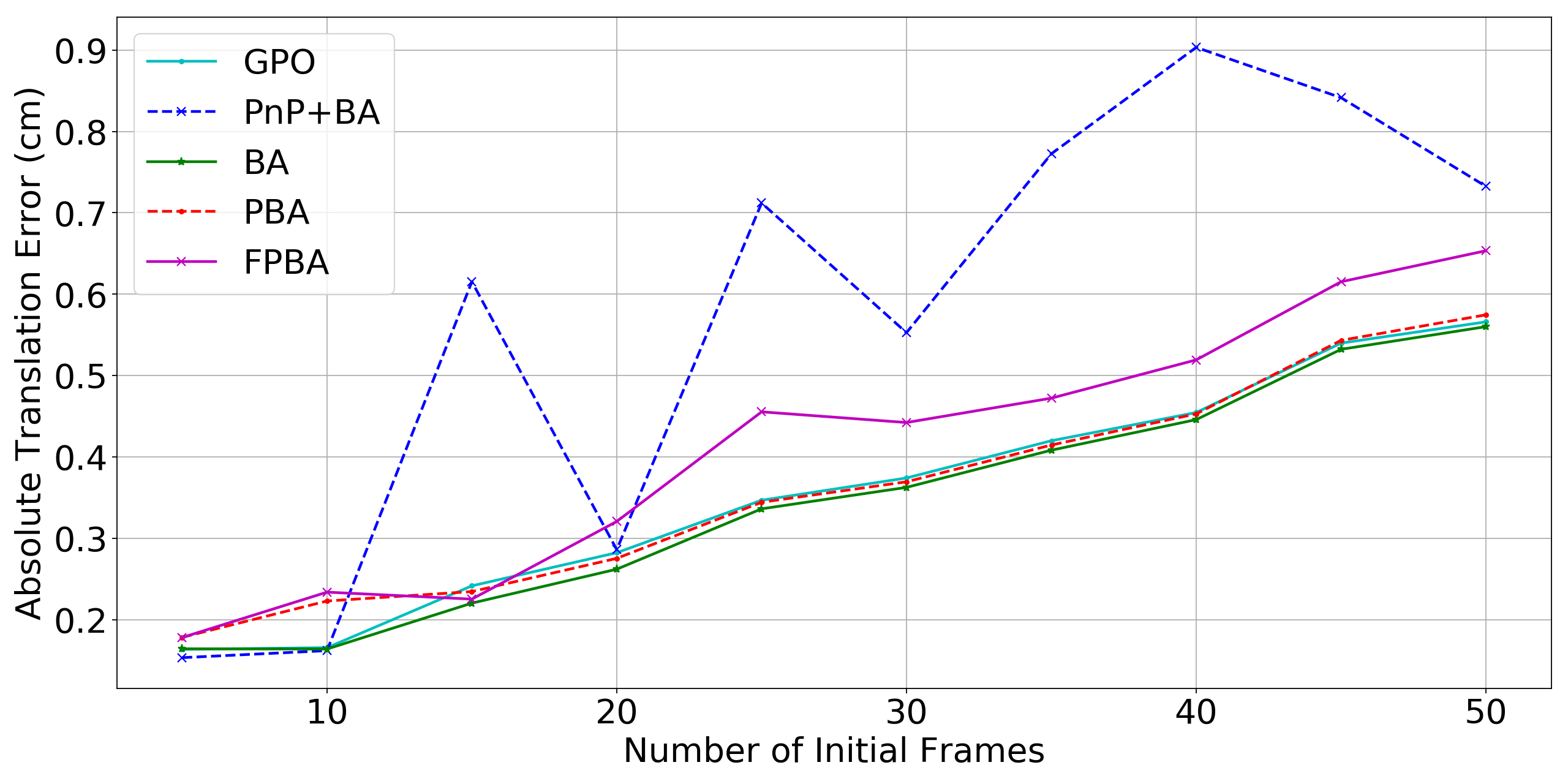}}
  \footnotesize\centerline{(a)}
\end{minipage}}
\vfill
\centerline{\begin{minipage}{0.9\linewidth}
  \centerline{\includegraphics[width=1.0\linewidth]{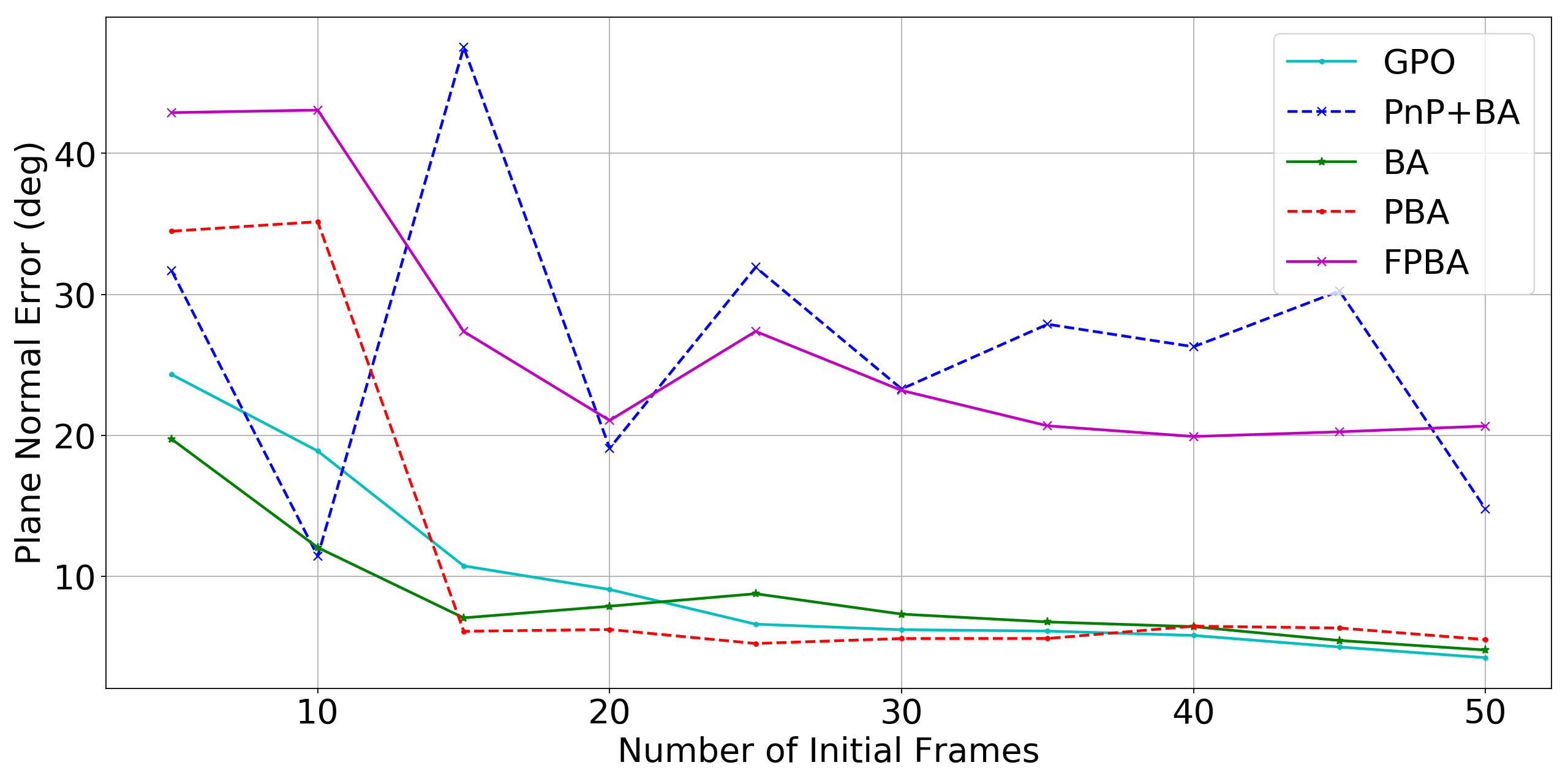}}
  \footnotesize\centerline{(b)}
\end{minipage}}
\vfill
\centerline{\begin{minipage}{0.9\linewidth}
  \centerline{\includegraphics[width=1.0\linewidth]{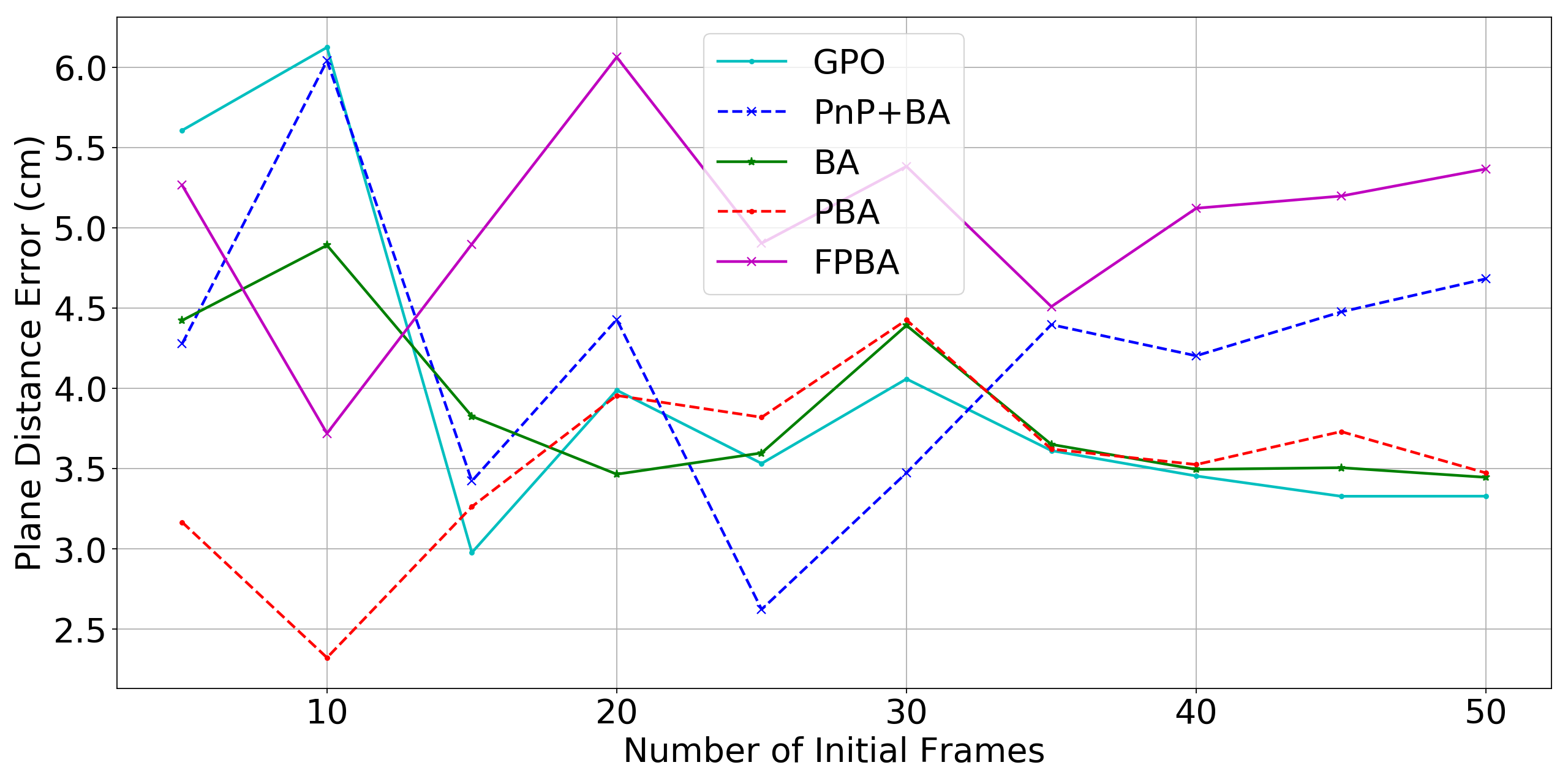}}
  \footnotesize\centerline{(c)}
\end{minipage}}
\caption{ATE, PNE and PDE of our GPO method and optimization-based baselines.}
\label{fig_ba}
\end{figure}

We compare our proposed GPO with the optimization-based algorithms in Fig. \ref{fig_ba}. GPO shows superior performance than BA with PnP \cite{qin_vins-mono:_2018} and FPBA methods. And it achieves significant better normal accuracy than PBA for fewer frames and performs the best among all methods when using more than 40 frames. We also find that our tuned BA with DBSCAN initialization has smaller errors than GPO for fewer frames, This may because it can make use of more points out of the planes to improve the pose estimation.

For BA-based methods, BA with clustering is better than that with PnP because more frame information is involved. Vanilla BA and plane-based BA performs better than FPBA, for the fixed plane normal given by DBSCAN is not exactly and it may cause a inaccurate direction during optimization process.

In addition, we find that the normal errors decrease with more frames. But for ATE, the errors increase with more frames due to the increasing alignment errors in evaluation. So the ATE metric may not correctly reveals the performance of methods for less frames.

We include a running time comparison in Table. \ref{tab_time} for 30 frame initialization. Our GPO runs fastest among the optimization methods for both the average time and the optimization time. This is because it has the fewest unknowns and constraints (see \ref{param} for details) and can be initialized without homography decomposition.

\newcommand{\tabincell}[2]{\begin{tabular}{@{}#1@{}}#2\end{tabular}}
\renewcommand\arraystretch{1.1}
\begin{table}[tb]
\caption{Average and optimization running time for different initialization methods with 30 frames.}
\begin{center}
\begin{tabular}{c|c|c|c|c|c}
 & \textbf{GPO} & \textbf{PnP + BA} & \textbf{BA} & \textbf{PBA} & \textbf{FPBA} \\
\hline
Avg time (ms) & \textbf{4.29} & 6.91 & 5.33 & 5.34 & 4.82 \\
\hline
Optim time (ms) & \textbf{25.84} & 89.14 & 48.47 & 80.99 & 34.22 \\
\end{tabular}
\label{tab_time}
\end{center}
\end{table}

\subsection{Qualitative Analysis}
\begin{figure}[tb]
  \centerline{\includegraphics[width=1.0\linewidth]{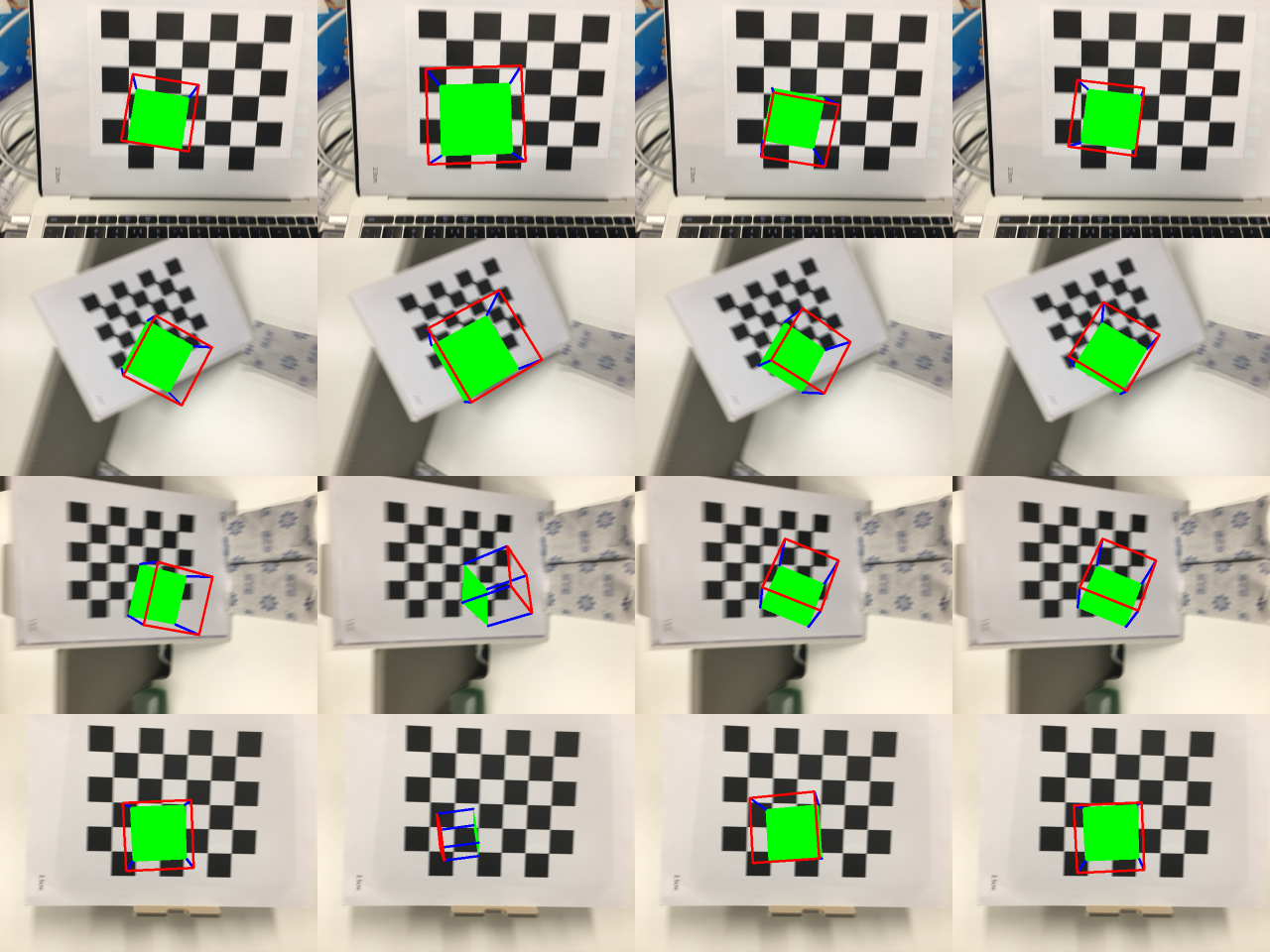}}
\caption{Visualization of initialization results with a green bottom cube. From left to right are results of GPO, PnP + BA, BA and PBA.}
\label{vis}
\end{figure}

Fig. \ref{vis} shows the qualitative results of different initialization methods with a green bottom cube. GPO, as shown in the first column yields best results, the orientation of cube is most consistent with the chessboard. PnP + BA, as shown in the second column, has the worst performance because it lacks the planar constraints. Interestingly, vanilla BA, as shown in the third column, outperforms PnP + BA mainly because the accumulated error from PnP and triangulation is not present. However, its performance is still inferior to the PBA as shown in the fourth column. 

\section{Conclusions}
In this paper, we present a novel initialization method GPO for monocular SLAM based on planar features. By combining multi-frame planar information, our method avoids the burdens of triangulation and homography decomposition. We validate the performance of our system on our chessboard datasets. The experimental results demonstrate that the performance of the proposed method is better than the baseline methods both on accuracy and time efficiency.


%%%%%%%%%%%%%%%%%%%%%%%%%%%%%%%%%%%%%%%%%%%%%%%%%%%%%%%%%%%%%%%%%%%%%%%%%%%%%%%%

%%%%%%%%%%%%%%%%%%%%%%%%%%%%%%%%%%%%%%%%%%%%%%%%%%%%%%%%%%%%%%%%%%%%%%%%%%%%%%%%
% \renewcommand*{\bibfont}{\footnotesize}
% \printbibliography
%\bibliographystyle{IEEEtran}
{\small
\bibliographystyle{IEEEtran}
\bibliography{total}
}

\section*{Appendices}

In this appendix, we present two extra experiments. 

\begin{figure}[htb]
\centerline{\begin{minipage}{0.9\linewidth}
  \centerline{\includegraphics[width=1.0\linewidth]{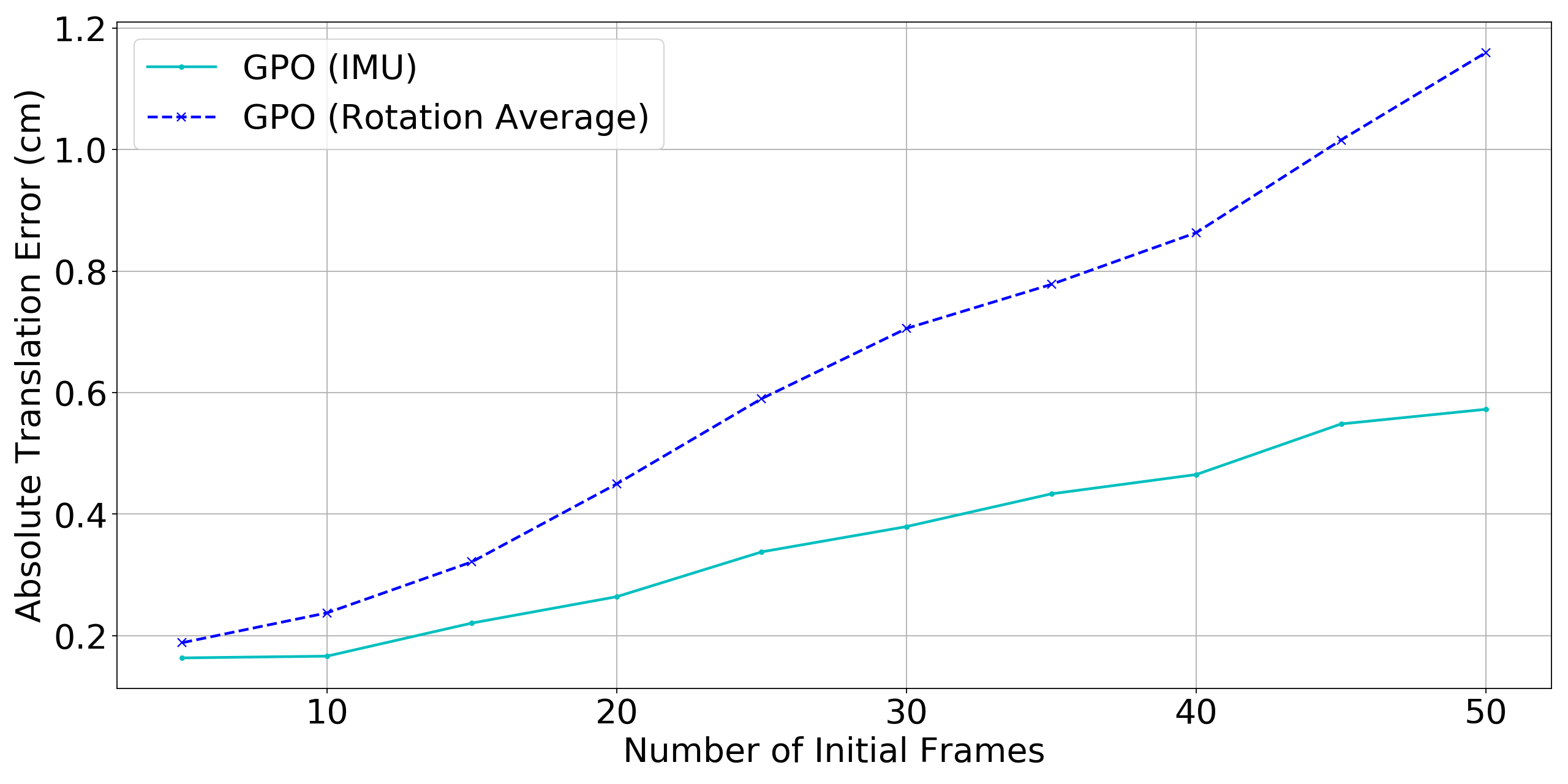}}
  \footnotesize\centerline{(a)}
\end{minipage}}
\vfill
\centerline{\begin{minipage}{0.9\linewidth}
  \centerline{\includegraphics[width=1.0\linewidth]{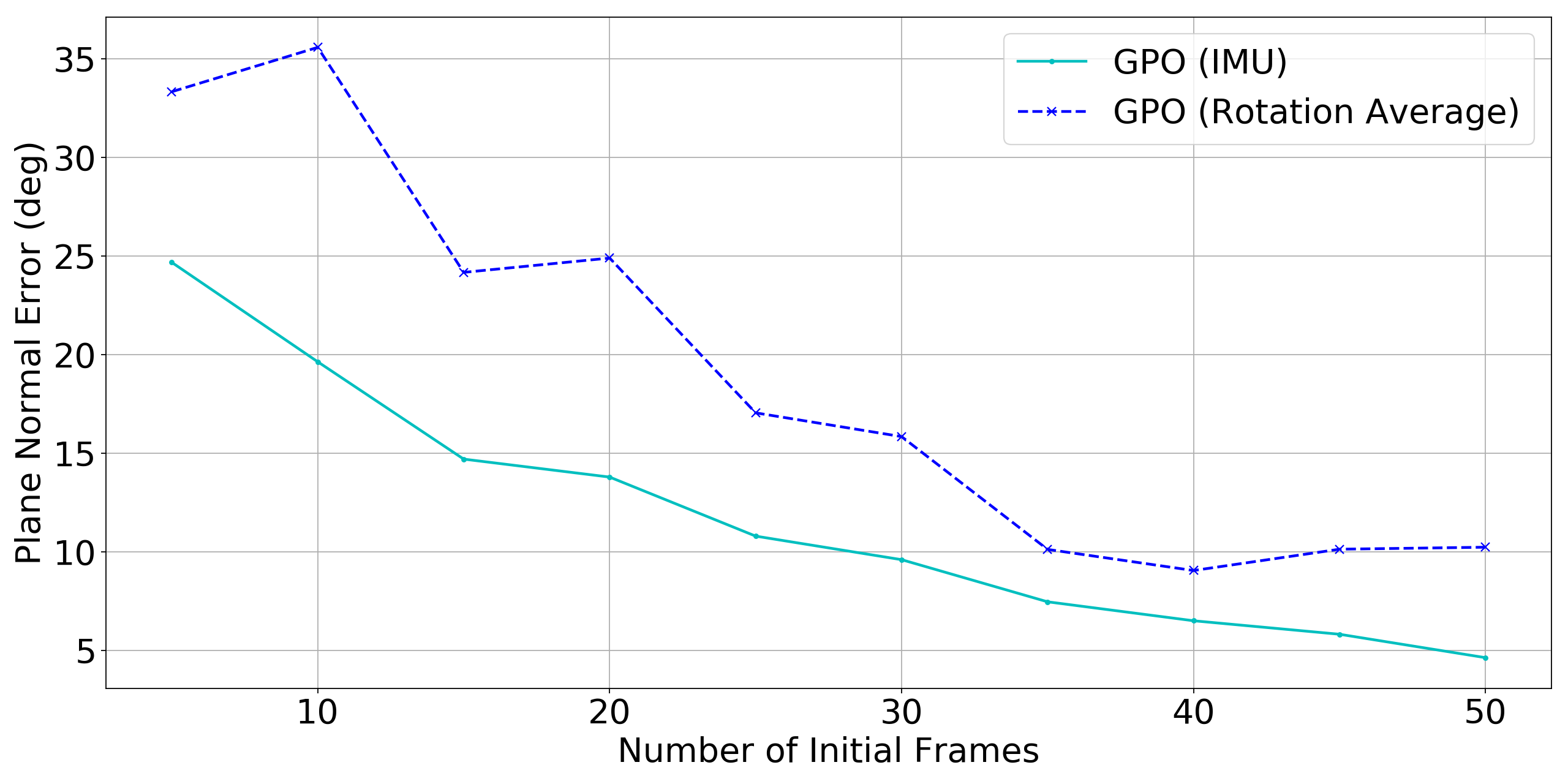}}
  \footnotesize\centerline{(b)}
\end{minipage}}
\vfill
\centerline{\begin{minipage}{0.9\linewidth}
  \centerline{\includegraphics[width=1.0\linewidth]{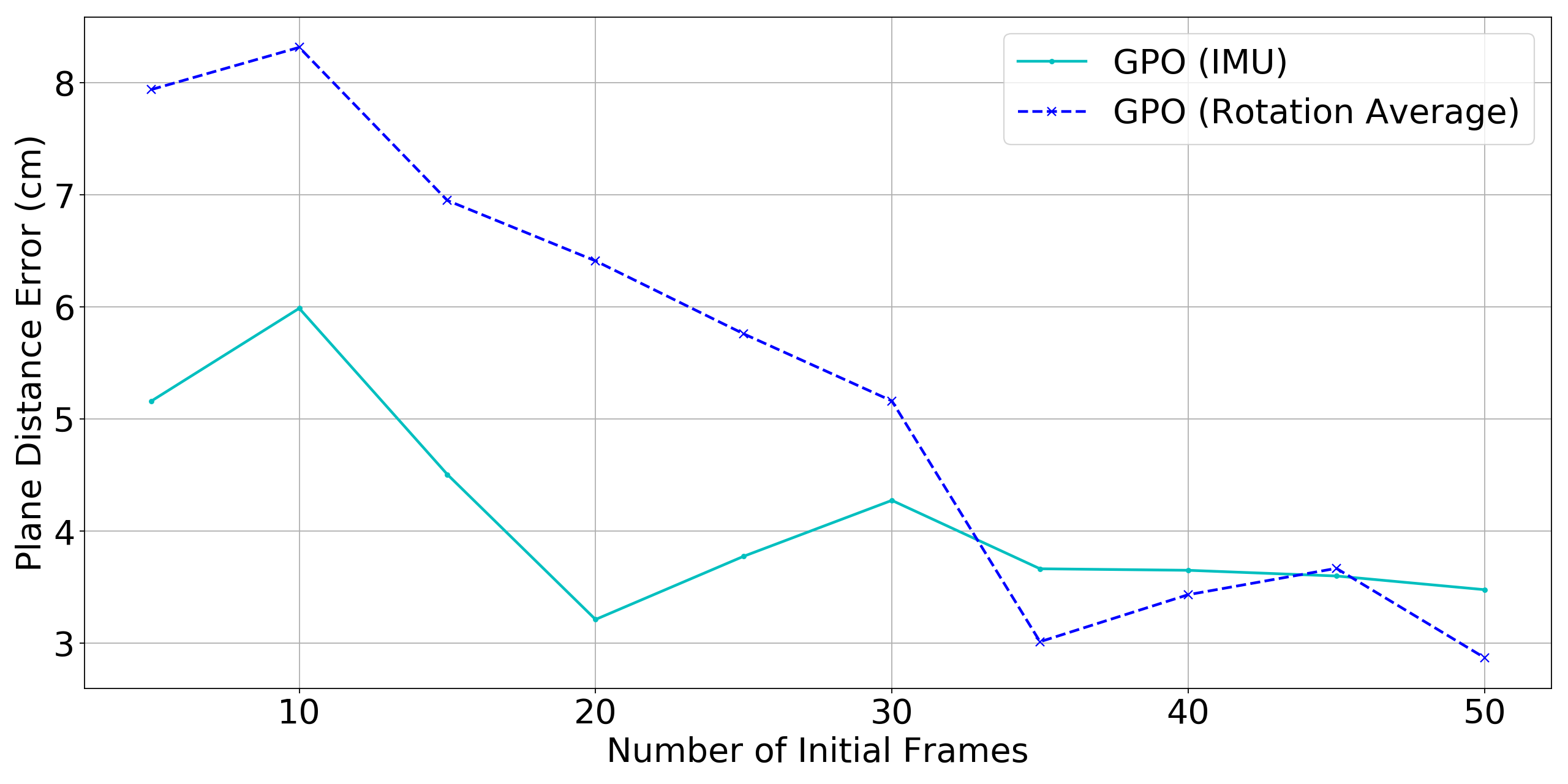}}
  \footnotesize\centerline{(c)}
\end{minipage}}
\caption{ATE, PNE and PDE of our GPO method with the rotations from IMU or rotation average.}
\label{fig_GPOAVR}
\end{figure}

The first experiment is the comparison of GPO using different rotation sources. The rotations from IMU are used in our main experiments. Here we test the results for GPO with rotations from a mature rotation average algorithm \cite{chatterjee2013efficient}, whose inputs are the estimated relative rotations with the five-point algorithm \cite{nister_efficient_2004}. The results are shown in Fig. \ref{fig_GPOAVR}. We can find that GPO with IMU rotations has smaller errors than that with the computed rotations on all the metrics.

\begin{figure}[t]
\centerline{\begin{minipage}{0.9\linewidth}
  \centerline{\includegraphics[width=1.0\linewidth]{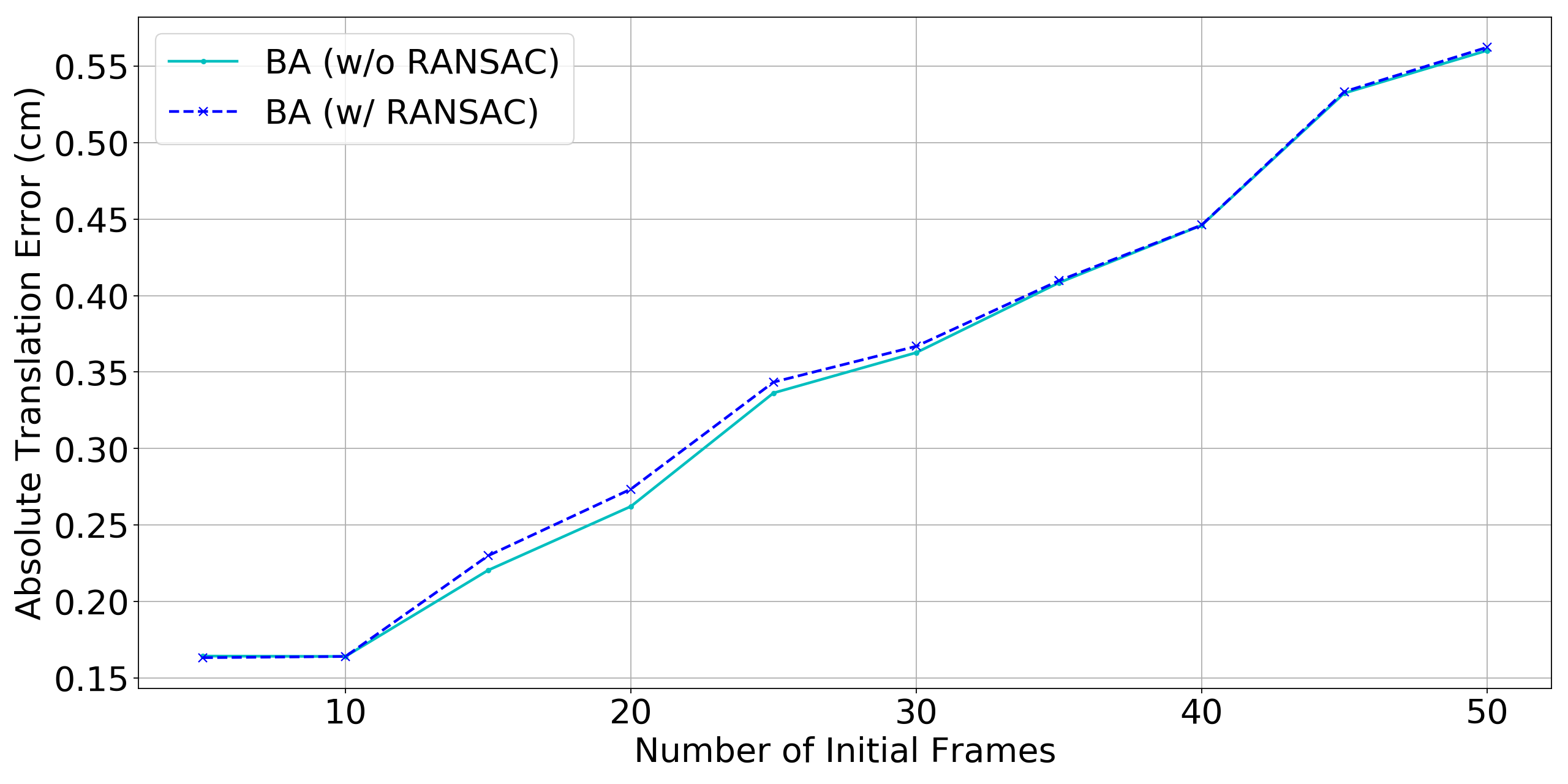}}
  \footnotesize\centerline{(a)}
\end{minipage}}
\vfill
\centerline{\begin{minipage}{0.9\linewidth}
  \centerline{\includegraphics[width=1.0\linewidth]{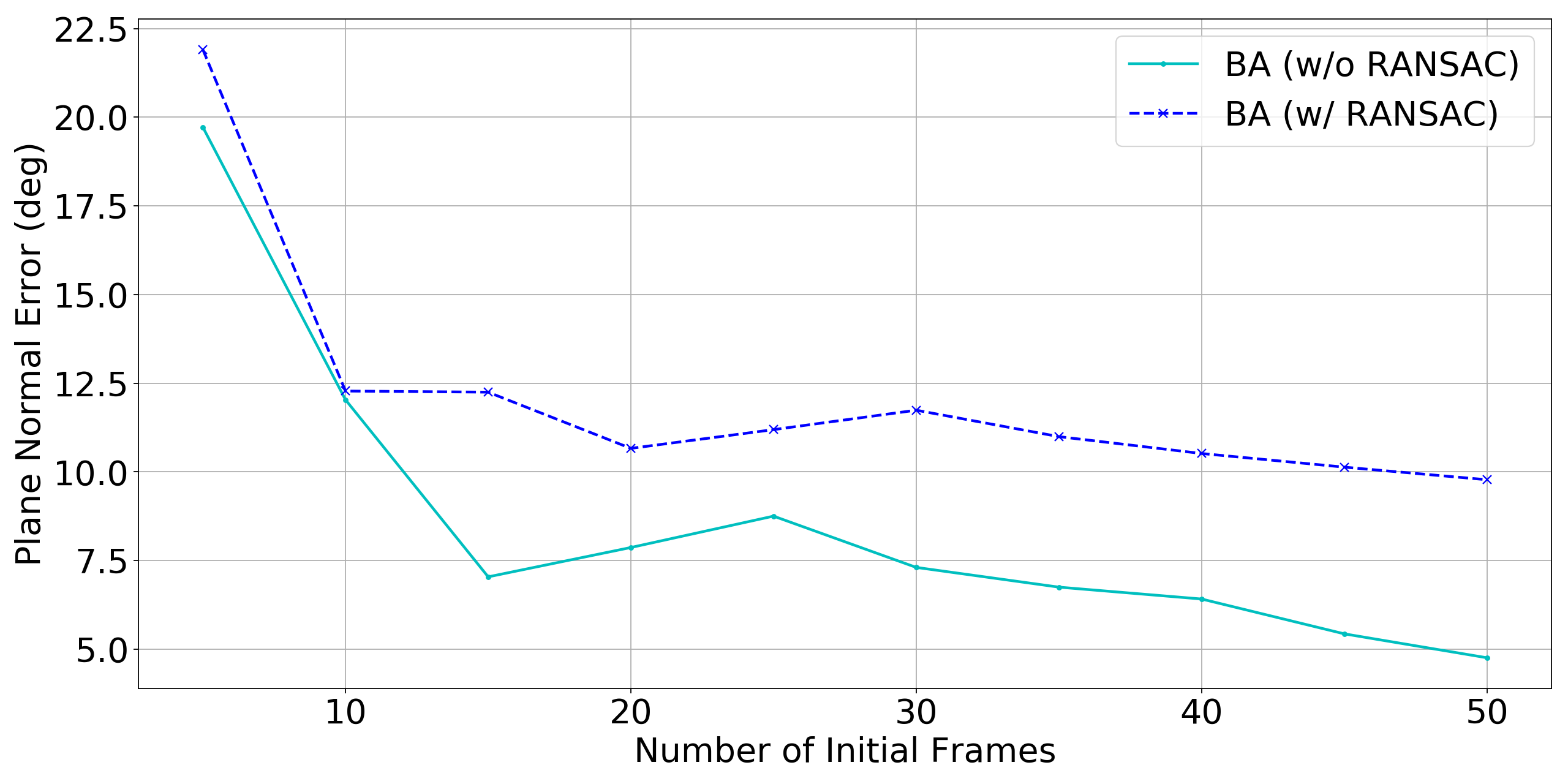}}
  \footnotesize\centerline{(b)}
\end{minipage}}
\vfill
\centerline{\begin{minipage}{0.9\linewidth}
  \centerline{\includegraphics[width=1.0\linewidth]{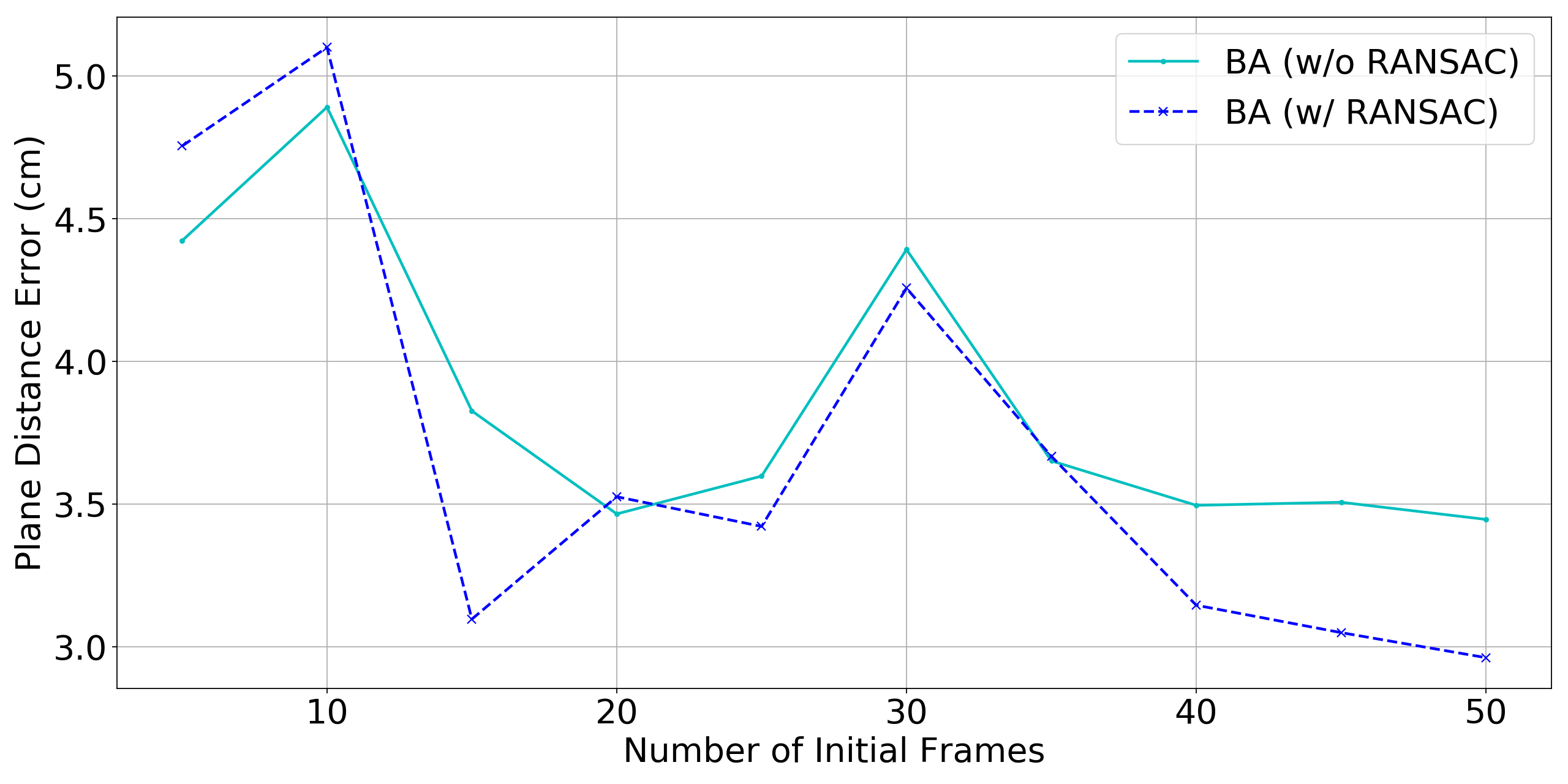}}
  \footnotesize\centerline{(c)}
\end{minipage}}
\caption{ATE, PNE and PDE of BA method with or without RANSAC.}
\label{fig_BARANSAC}
\end{figure}

The second experiment is about BA with or without homography RANSAC. BA does not need to force all the points on the same plane, so in our main experiments we do not apply the RANSAC for BA. Here we investigate the effects for the RANSAC. From the ATE in Fig. \ref{fig_BARANSAC}, these two methods perform comparably. As for the plane metrics, BA without RANSAC significantly outperforms the one with RANSAC on plane normal estimation. This may results from the reduction of the number of points after RANSAC, which leads to less constraints during the optimization process. For PDE, BA without RANSAC has larger errors than that with RANSAC. It may because BA with RANSAC has less map points out of the plane, making the average distance of the plane more accurate. 

%\addtolength{\textheight}{-12cm}

\end{document}